\documentclass[10pt]{article} 
\usepackage[preprint]{tmlr}

\usepackage[utf8]{inputenc} % allow utf-8 input
\usepackage[T1]{fontenc}    % use 8-bit T1 fonts

\usepackage{anyfontsize}
\usepackage{algorithm, algpseudocode, amsfonts, amsthm, amsmath, amssymb, appendix}

\usepackage{bm, bbm, booktabs}
\usepackage{color, comment}
\usepackage{graphicx}
\usepackage{enumitem, epstopdf, framed, hyperref}
\definecolor{darkblue}{rgb}{0.10, 0.20, 0.65}
\definecolor{darkred}{rgb}{0.70, 0.00, 0.00}
\definecolor{darkgreen}{rgb}{0.20, 0.50, 0.20}
\hypersetup{
  colorlinks=true,
  breaklinks=true,
  linkcolor=darkred,
  urlcolor=darkblue,
  anchorcolor=darkblue,
  citecolor=darkblue
}
\usepackage{latexsym, lineno}
\usepackage{mathrsfs, microtype, minitoc, multirow, multirow, nicefrac}
\usepackage{pdflscape, pifont}
\usepackage{silence}
\usepackage{subcaption}
\usepackage{tabularx, tcolorbox, threeparttable, txfonts}

\usepackage{url, utfsym}
\usepackage{wrapfig}

\def\1{\bm{1}}

\def\btheta{{\bm{\theta}}}
\def\bJ{{\bm{J}}}
\def\bomega{{\boldsymbol{\omega}}}

\DeclareMathAlphabet{\mathsfit}{\encodingdefault}{\sfdefault}{m}{sl}
\SetMathAlphabet{\mathsfit}{bold}{\encodingdefault}{\sfdefault}{bx}{n}

\newtheorem{theorem}{Theorem}

\newcounter{assumption}%
\renewcommand{\theassumption}{\arabic{assumption}}

\title{Exploring Non-Convex Discrete Energy Landscapes:\\ An Efficient Langevin-Like Sampler with Replica Exchange}

\author{\name Haoyang Zheng \email zheng528@purdue.edu \\
      \addr School of Mechanical Engineering\\
      Purdue University
      \AND
      \name Hengrong Du \email hengrond@uci.edu \\
      \addr Department of Mathematics \\
      University of California, Irvine
      \AND
      \name Ruqi Zhang \email ruqiz@purdue.edu\\
      \addr Department of Computer Science\\
      Purdue University
      \AND
      \name Guang Lin \email guanglin@purdue.edu\\
      \addr Department of Mathematics and School of Mechanical Engineering\\
      Purdue University}

\begin{document}

\maketitle

\begin{abstract}

Gradient-based Discrete Samplers (GDSs) are effective for sampling discrete energy landscapes. However, they often stagnate in complex, non-convex settings. To improve exploration, we introduce the Discrete Replica EXchangE Langevin (DREXEL) sampler and its variant with Adjusted Metropolis (DREAM). These samplers use two GDSs \citep{zhang2022langevin} at different temperatures and step sizes: one focuses on local exploitation, while the other explores broader energy landscapes. 
When energy differences are significant, sample swaps occur, which are determined by a mechanism tailored for discrete sampling to ensure detailed balance. Theoretically, we prove that the proposed samplers satisfy detailed balance and converge to the target distribution under mild conditions. Experiments across 2d synthetic simulations, sampling from Ising models and restricted Boltzmann machines, and training deep energy-based models further confirm their efficiency in exploring non-convex discrete energy landscapes.

\end{abstract}

\section{Introduction}\label{sec:intro}

Sampling from high-dimensional discrete distributions has been an important task for decades across applications in texts \citep{mikolov2013efficient, devlin2018bert}, images \citep{krizhevsky2012imagenet, ronneberger2015u}, signal processing \citep{mallat1989theory, donoho2006compressed}, genome sequences \citep{metzker2010sequencing, macosko2015highly}, etc. However, the exponential growth in the number of configurations makes sampling from $\pi(\btheta) \propto \exp\left[U(\btheta)\right]$ computationally prohibitive. The computational burden comes from evaluating the exact probabilities and normalizing constants, which makes exact sampling impossible in practice. Algorithms such as rejection sampling \citep{neumann1951various}, Swendsen-Wang \citep{swendsen1987nonuniversal}, and Hamze-Freitas \citep{hamze2004fields} leverage special structures within the problem to make global updates. In more general settings, these methods may suffer from slow exploration, local dependencies, and poor convergence.

% \begin{wrapfigure}{r}{0.39\textwidth}
%    \begin{center}
%    \vskip -0.5in
%      \includegraphics[width=0.37\textwidth]{figs/framework.pdf}
%    \end{center}
%    \vskip -0.18in
%    \caption{{\small DREXEL \& DREAM sample trajectory in discrete domains. \textcolor{darkblue}{Blue denotes a low-temperature sampler}, and \textcolor{red}{red high-temperature sampler}. They exchange samples following a swap mechanism.}}
%    \label{fig:framework}
%    \vspace{-0.40 in}
% \end{wrapfigure}
To make high-dimensional discrete sampling more efficient, Locally Balanced Proposals (LBPs) \citep{zanella2020informed, sun2021path} improved acceptance rates by adjusting proposal distributions based on the likelihood ratio. Early LBPs updated one coordinate at a time \citep{zanella2020informed, grathwohl2021oops}, and \citet{grathwohl2021oops} developed gradient-based discrete sampler (GDS) to update coordinately. Later, \citet{zhang2022langevin} further extended GDSs by updating all coordinates simultaneously, which enhances efficiency and scalability for large-scale, high-dimensional computations on GPUs and TPUs.

% a locally-balanced proposal (LBP) \citep{zanella2020informed} was introduced. These samplers use ratio-informed proposal distributions \citep{grathwohl2021oops, sun2021path} that adjust proposals based on the likelihood ratio between the current and proposed states, which improves acceptance rates and overall efficiency \citep{goshvadi2024discs}. Gradient-based Discrete Samplers (GDSs) \citep{zhang2022langevin} further improve upon LBPs by utilizing gradient updates across all coordinates, which makes it more efficient and scalable for large-scale, high-dimensional computations on GPUs and TPUs.\ruqi{zanella2020informed and sun2021path are LBP. grathwohl2021oops and zhang2022langevin are LBP with gradient. zanella2020informed and grathwohl2021oops update one coordinate. zhang2022langevin and sun2021path update multiple coordinates. might mention in this order: LBP, then gradient-based discrete sampler, then Langevin-like}

\begin{wrapfigure}{r}{0.38\textwidth}
   \begin{center}
   \vskip -0.05in
     \includegraphics[width=0.38\textwidth]{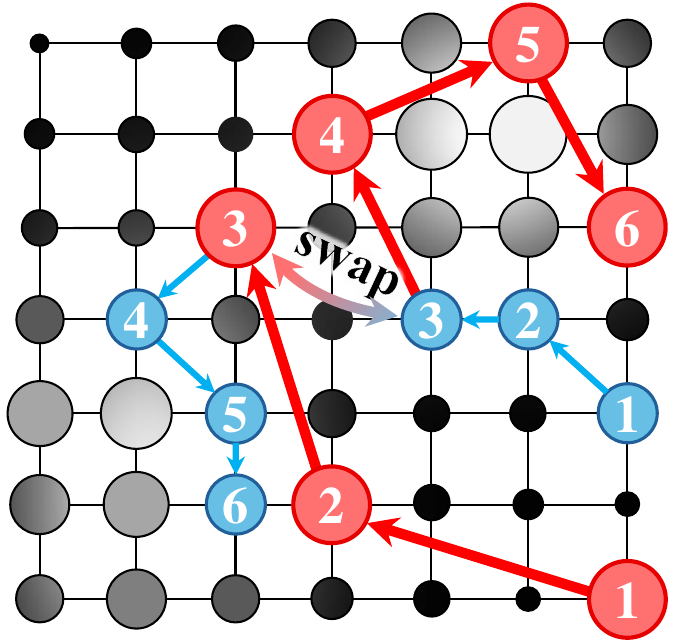}
   \end{center}
   \vskip -0.18in
   \caption{{\footnotesize DREXEL \& DREAM sample trajectory in discrete domains. \textcolor{darkblue}{Blue denotes a low-temperature sampler}, and \textcolor{red}{red high-temperature sampler}. They exchange samples following a swap mechanism.}}
   \label{fig:framework}
   \vspace{-0.20 in}
\end{wrapfigure}
Despite improvements in LBPs, how to balance the trade-off between ``global exploration'' and ``local exploitation'' remains a challenge. High-dimensional discrete distributions are highly multi-modal, with deep and narrow wells caused by intrinsic discontinuities. Gradient-based LBPs, although effective, tend to get trapped in local modes due to their reliance on local gradients and small noise, which is insufficient for escaping these traps.
% because the reliance on local gradient information and small noise perturbations is insufficient to escape these traps.

% \haoyang{explain the motivation}

To bridge this gap, we propose two samplers: Discrete Replica EXchangE Langevin (DREXEL) and Discrete Replica Exchange with Adjusted Metropolis (DREAM). These samplers combine GDS with the replica exchange Markov Chain Monte Carlo (reMCMC) \citep{chen2019accelerating} for efficient exploration of non-convex discrete spaces. As illustrated in Figure \ref{fig:framework}, the samplers employ two GDSs at different temperatures and step sizes: the low-temperature sampler focuses on local exploitation, while the high-temperature sampler escapes local traps for broader exploration. Sample swaps occur when energy differences are significant, governed by a mechanism tailored for discrete sampling to ensure detailed balance.
% Sample exchanges occur when the temperature difference is significant. 
The combination of replica exchange and GDS makes
% for high-dimensional inference and scalable optimization \ruqi{not clear what this sentence means}, making 
them particularly effective for sampling from complex discrete structures in modern applications. The primary contributions in this work are summarized as follows:
\begin{tcolorbox}[frame empty, left = 0.0mm, right = 0.0mm, top = 0.0mm, bottom = 0.0mm, width=\textwidth]
\begin{itemize}[leftmargin=10 pt]
    \item A novel integration of GDS with replica exchange to improve non-convex exploration;%\ruqi{to achieve what}
    \item A swap mechanism tailored for detailed balance and sample efficiency in discrete sampling; % for efficient discrete sampling \ruqi{how tailored? might give a bit more detail}\ruqi{to enable what};
    \item Theoretical analysis of improved mixing rates over na\"ive discrete Langevin-like samplers; % \ruqi{enhanced rate compared to what}
    \item Superior performance in synthetic tasks, Ising models, restricted Boltzmann machines, and energy-based deep learning models for navigating non-convex discrete energy landscapes. % \ruqi{may mention the specific tasks that we tested and how the results are improved if they are significant}
\end{itemize}
\end{tcolorbox}
\vspace{-0.05 in}

\section{Related Work}\label{sec:related}
% \ruqi{i'd suggest changing the title to "Gradient-based Discrete Sampling". since the key point is to emphasize the use of gradient, not locally-balanced. }

\textbf{Gradient-based Discrete Sampling} becomes popular for complex discrete sampling tasks and its original idea comes from LBPs.
% \ruqi{this is not true. efficiency can also come from gradient info}
The concept of LBPs, as introduced by \citet{zanella2020informed}, utilized local information in the form of density ratios to improve the sample efficiency.
% integrates Metropolis-Hastings (MH) into its design. \ruqi{this is not what LBP proposes}. 
\citet{grathwohl2021oops} expanded LBPs by the use of first-order Taylor approximation, which further ensures computational feasibility. To improve sampling in high-dimensional discrete spaces, LBPs were extended to cover larger neighborhoods by performing a sequence of small moves \citep{sun2021path}. \cite{zhang2022langevin} further developed GDSs, which adapt the continuous Langevin MCMC to discrete spaces and allow parallel updates of all coordinates based on gradient information. Subsequently, GDSs were improved through the introduction of an adaptive mechanism by which the step size can be automatically adjusted \citep{sun2023any}. Most recently, \cite{pynadath2024gradient} introduced an automatic cyclical scheduling approach in step sizes to better handle multi-modal distributions by alternating between exploration and exploitation phases. While these methods have shown promise, sampling from highly non-convex discrete distributions remains challenging, particularly when dealing with strongly correlated variables or energy-based deep learning models. 

% \ruqi{it might be better to change this paragraph to "Replica Exchange MCMC" which is the most related literature to this paper?}

\textbf{Replica Exchange} MCMC is a powerful method that enhances exploration in complex, multi-modal distributions, and a variety of related algorithms build on this. For instance, unadjusted Langevin MCMC \citep{durmus2017nonasymptotic} leverages gradient information to guide proposals but lacks the exchange mechanism. Importance sampling \citep{wang2001efficient} adjusts for the discrepancy between target and proposal distributions, which offers flexibility in sampling but without temperature-based exchanges. Simulated tempering \citep{lee2018beyond} further refined the temperature-scaling strategy by dynamically adjusting the temperature of a single chain. Recently, \citet{zhangcyclical} proposed a cyclical step-size scheduler to maintain a balance between exploration and exploitation. To enhance the exploration, reMCMC runs multiple chains at different temperatures and allows for chain swaps between them. \citet{dong2022spectral} analyzed its mixing by quantifying the spectral gap, and \citet{deng2020non, deng2022accelerating} validated its efficiency in large-scale deep learning tasks. Despite its success in continuous sampling and high-dimensional settings, to the best of our knowledge, its potential has not been studied in sampling from discrete distributions.

% further mitigates these challenges by employing multiple temperatures and state-swapping to achieve faster mixing rates. Its efficacy is further supported by both theory \citet{dong2022spectral} and experiments \citep{deng2020non, deng2022accelerating}.\ruqi{are there works on replica exchange in discrete spaces?}

% \ruqi{this paragraph seems not very related to this paper. may consider removing it or put in appendix}

\section{Preliminaries}\label{sec:prelim}
\textbf{The Target Distribution} $\pi:=\Theta\rightarrow [0, 1]$ denotes the probabilistic model we are sampling from:
\begin{equation}\label{eq:target}
    \pi(\btheta) =\frac{1}{Z} \exp\left[\frac{U(\btheta)}{\tau}\right],\quad \forall \btheta \in \Theta,\ \  \Theta\subset \mathbb R^{\mathbf{d}}.
\end{equation}
Here $\btheta$ 
% \ruqi{why do we have two variables? should the target distribution be just $\pi(\theta)$?}
is $\mathbf{d}$-dimensional variable, $\tau=1.0$ denotes the respective temperature, $\Theta$ is a finite domain, $U$ represents the energy function, and $Z$  normalizes the distribution. Following the traditional settings in discrete sampling, we assume: 
\begin{tcolorbox}[frame empty, left = 0.0mm, right = 0.0mm, top = 0.0mm, bottom = 0.0mm]
1. The sampling domain is coordinate-wisely factorized where $\Theta = \prod_{d=1}^\mathbf{d} \Theta_d$, and we primarily consider the binary cases $\Theta = \{0,1\}^\mathbf{d}$ or categorical $\{0,1,\dots, N-1\}^\mathbf{d}$; 

2. The energy function is differentiable across $\mathbb{R}^\mathbf{d}$. 
\end{tcolorbox}
The primary goal is to design an efficient sampler to approximate $\pi$ within a finite sample size. The empirical distribution derived from these samples converges to $\pi$, with the approximation error bounded by a constant $\epsilon > 0$ under specific metrics.

\textbf{Replica Exchange} MCMC is a popular sampling method for non-convex exploration in continuous spaces. It updates according to the following dynamics:
\begin{equation}\label{eq:langevin_chain}
    \btheta_{i+1}^{(k)} = \btheta_i^{(k)} + \frac{\alpha_k}{2}\nabla U(\btheta_i^{(k)}) + \sqrt{\alpha_k\tau_k}\xi_k,\ \ k=1,2; \ \ i=1, 2, \cdots, I.
\end{equation}
Here $\alpha_1, \alpha_2\in\mathbb R^+$ represent step sizes, $\tau_1, \tau_2\in\mathbb R^+$ are temperatures, and $\xi_1, \xi_2$ are independent Gaussian noises drawn from $\mathcal{N}\left(\boldsymbol 0,\mathbf{I}_{\mathbf{d}\times \mathbf{d}}\right)$. The typical setup assumes $\alpha_1 < \alpha_2$ and $\tau_1 < \tau_2$, with the first chain in \eqref{eq:langevin_chain} labeled as the low-temperature chain and the second as the high-temperature chain. The gradient $\nabla U(\cdot)$ directs the algorithm toward high-probability regions. To further improve the mixing rate over Langevin MCMC, reMCMC facilitates interaction via a chain swap mechanism. Specifically, the probability to swap the $i$-th samples between $\btheta_i^{(1)}$ and $\btheta_i^{(2)}$ is determined by $\rho\min\left\{1,\  S(\btheta_i^{(1)},\btheta_i^{(2)})\right\}$. The swap intensity is regulated by $\rho > 0$, and {the swap function} $S:=\Theta \times \Theta\rightarrow \mathbb R^+$ is given as follows:
\begin{equation}\label{eq:swap_naive}
\begin{split}
    S(\btheta^{(1)}, \btheta^{(2)})= e^{ \left(\frac{1}{\tau_2}-\frac{1}{\tau_1}\right)\left[U(\btheta^{(1)})-U(\btheta^{(2)})\right]}.
\end{split}
\end{equation}
Intuitively, the swap probability in reMCMC depends on the energy estimated at $\btheta^{(1)}$ and $\btheta^{(2)}$. When the low-temperature chain gets stuck in a local minimum and the high-temperature chain escapes to find modes with significantly lower energy, the chain will swap their samples with high probability. This enables the low-temperature chain to better characterize the newly discovered modes, while the high-temperature chain continues to search across the energy landscape. As mentioned in \citet{chen2019accelerating}, reMCMC behaves as a reversible Markov jump process due to its swap mechanism, which converges to a similar invariant distribution in \eqref{eq:target} while the parameters can explore over $\mathbb R^{\mathbf{d}}$.

While reMCMC is a powerful tool for non-convex exploration, its update may fail to preserve the target distribution due to discretization errors introduced by the finite step size \citep{welling2011bayesian}. According to \citet{roberts1996exponential}, selecting an inappropriate step size can lead to a transient Markov chain without a stationary distribution. To mitigate such bias, two main approaches are commonly used: decaying step sizes \citep{vollmer2016exploration, teh2016consistency} and Metropolis-Hastings (MH) corrections \citep{dwivedi2019log, chewi2021optimal}. While implementing decaying step sizes is straightforward and does not require additional computational burden, the second approach is more favorable due to its specific advantages in discrete sampling, which will be elaborated on later.

\textbf{Metropolis-Hastings Correction} is considered to correct discretization errors and ensure convergence to the target distribution. Specifically, at each iteration, a new candidate $\bomega\leftarrow \btheta_{i+1}$\footnote{For clarity and conciseness, we omit the chain index when there is no need to specify it.} is first generated with \eqref{eq:langevin_chain}. To ensure that the resulting samples come from the target distribution, the MH step determines whether to accept or reject the candidate with $\mathcal{A}:=\Theta\times\Theta\rightarrow [0, 1]$:
\begin{equation}\label{eq:mh_langevin}%{eq:mh}
    {\mathcal{A}(\bomega, \btheta_i)} = \min \left\{1, \frac{\pi(\bomega) q\left(\btheta_i \mid \bomega\right)}{\pi(\btheta_i) q\left(\bomega \mid\btheta_i\right)} \right\},
\end{equation}
where $q:=\Theta\times\Theta\rightarrow [0, 1]$ is the transition probability mapping from the current sample $\btheta_i$ to the next sample $\btheta_{i+1}$. 
% Following \citet{dwivedi2019log}, this acceptance probability can be simplified for LAs as follows:
% \begin{equation}\label{eq:mh_langevin}
%     {\mathcal{A}}(\bomega, \btheta_i) = \min \left\{ 1, \frac{\exp\left( U(\btheta_i) + \|\bomega - \btheta_{i} - \frac{\alpha}{2\tau}\nabla U(\btheta_i)\|^2 / 4\alpha \right)}{\exp\left(U(\bomega)+\|\btheta_{i} - \bomega - \frac{\alpha}{2\tau}\nabla U(\bomega)\|^2 / 4\alpha\right)} \right\}.
% \end{equation}
With probability ${\mathcal{A}}(\bomega, \btheta_i)$, the candidate $\bomega$ is accepted in the current step; otherwise, it retains the current $\btheta_i$. This adjustment preserves the correct stationary distribution. Furthermore, because Langevin MCMC allows each sample to access any point in $\mathbb{R}^{\mathbf d}$, it further ensures the Markov chain is both irreducible and ergodic \citep{diaconis1997markov, meyn2012markov}.
% \ruqi{I suggest to put Sec 4.1 under preliminary, to make it clear that it is existing work, not sth we propose}

% \subsection{}
\textbf{Discrete Langevin Sampler} (DLS)
% \ruqi{this section can be shortened if lack of space} 
is a gradient-based approach for sampling from high-dimensional discrete distributions. Inspired by Langevin MCMC, DLS updates all coordinates in parallel from a single gradient computation to function effectively in discrete settings. Specifically, for a target distribution $\pi\propto\exp\left[U(\cdot)\right]$, DLS generates a new sample $\btheta_{i+1}$ inspired by the Taylor expansion:
\begin{equation}\label{eq:dlp}
    q\left(\btheta_{i+1} \mid \btheta_{i}\right) = \frac{\exp \left( -\frac{1}{2 \alpha} \left\|\btheta_{i+1} - \btheta_{i} - \frac{\alpha}{2\tau} \nabla U(\btheta_{i})\right\|_2^2 \right)}{Z_\Theta(\btheta_i)}, \ \ \btheta_{i}, \btheta_{i+1}\in\Theta.
\end{equation}
Here $\nabla U(\btheta)$ is the gradient of the energy function evaluated at $\btheta$, and $Z_\Theta(\btheta)$ normalizes the distribution:
\begin{equation}\label{eq:normalize_trans}
    Z_\Theta(\btheta_i) = \sum_{\btheta_{i+1}\in\Theta}\exp\left(-\frac{1}{2 \alpha} \left\|\btheta_{i+1} - \btheta_{i} - \frac{\alpha}{2\tau} \nabla U(\btheta_{i})\right\|_2^2\right).
\end{equation}
This proposal distribution allows DLS to make larger, parallel updates while maintaining computational efficiency. As the dimension $\mathbf d$ in parameter space grows, the cost of computing \eqref{eq:normalize_trans} becomes prohibitively expensive. A key insight is that the update rule can be factorized by coordinate:
\begin{equation}\label{eq:categorical}
    \theta_{i+1,d}\sim\text{Categorical}\left[\text{Softmax} \left(\frac{1}{2\tau}\nabla U(\btheta_i)_d(\theta_{i+1,d}-\theta_{i,d}) - \frac{(\theta_{i+1,d}-\theta_{i, d})^2}{2\alpha}\right)\right],
\end{equation}
where $d = 1, 2, \dots, \mathbf{d}$ is the dimension index, and $\theta_{i,d}$ represents the $i$-th sample in the $d$-th dimension. This algebraic expansion, following the binomial theorem, works because $\left(\nabla U(\btheta_i)_d\right)^2$ is independent of $\theta_{i+1,d}$. It makes DLS scalable and computationally efficient for complex distributions \citep{zhang2022langevin}. Furthermore, the first term in \eqref{eq:categorical} biases the proposal towards low-energy regions, where the gradient points towards increasing probability; the second term acts as a regularizing factor, which penalizes large jumps unless they are strongly favored by the gradient.

DLS can operate with or without MH corrections. Without corrections, it is simplified to the Discrete Unadjusted Langevin Algorithm (DULA), which is computationally efficient but may introduce bias. With corrections in \eqref{eq:mh_langevin}, it becomes the Discrete Metropolis-adjusted Langevin algorithm (DMALA), which corrects bias at an increased computational cost. Both DULA and DMALA employ non-local proposals specifically for the heat kernel to enable more efficient sampling \citep{sun2023any}. %Subsequently, these samplers are extended to enhance the exploration of discrete energy landscapes.

% % \clearpage
\section{Discrete Langevin Sampler with Replica Exchange}\label{sec:discrete}
The proposed DLS variants are present here, which incorporate replica exchange and a customized sampler swap mechanism to ensure detailed balance. The complete algorithm is provided at the end.

\subsection{Discrete Samplers with Different Temperatures}
A key challenge with the na\"ive DLS is the tendency to become trapped in local modes, particularly in non-convex landscapes. To mitigate this, we introduce DREXEL, which incorporates replica exchange to enable efficient exploration across different local modes. Specifically, we employ two samplers separately with distinct step sizes and temperatures to approximate the target distribution:
\begin{equation}\label{eq:replica_discrete}
    \begin{split}
        \text{Categorical}\left[\text{Softmax}\left(\frac{1}{2\tau_k}\nabla U(\btheta_i^{(k)})_d\left(\theta_{i+1,d}^{(k)}-\theta_{i,d}^{(k)}\right) - \frac{\left(\theta_{i+1,d}^{(k)}-\theta_{i, d}^{(k)}\right)^2}{2\alpha_k}\right)\right], \quad k=1,2.
    \end{split}
\end{equation}
Here $\tau_1<\tau_2$ and $\alpha_1<\alpha_2$, with $k=1$ being the low-temperature and $k=2$ the high-temperature sampler. Intuitively, larger step sizes and higher temperatures encourage more exploratory moves, which allows the sampler to escape local modes through non-local jumps and explore different regions of the energy landscapes. This, on the downside, raises the rejection rate, as large jumps often land in low-probability regions, and introduce additional bias when approximating the target distribution.

To mitigate the bias, we further propose DREAM, which incorporates MH steps post-generation of new samples. Once the new samples are produced through \eqref{eq:replica_discrete}, the acceptance rates ${\mathcal{A}}(\btheta_{i+1}^{(1)}, \btheta_i^{(1)})$ and ${\mathcal{A}}(\btheta_{i+1}^{(2)}, \btheta_i^{(2)})$ are estimated with \eqref{eq:mh_langevin}. The new samples are accepted with probability ${\mathcal{A}}$ or rejected with $1-{\mathcal{A}}$. The acceptance rates of two samplers are independent of one another. While the high-temperature sampler typically exhibits a lower acceptance rate than the low-temperature one, the rejection mechanism ensures that both samplers in DREAM converge to the target asymptotically.

% \haoyang{use samples or proposals or states?} 
It should be noted that while decaying step sizes are commonly advantageous in Langevin MCMC for handling big data \citep{teh2016consistency}, they present potential challenges in discrete sampling. In discrete spaces, small steps do not equate to gradual movements as they do in continuous spaces. Instead, they tend to repeatedly propose nearly identical samples, which causes the sampler to become trapped in local regions. This problem becomes severe when dealing with non-convex energy landscapes, where a decaying step size worsens the issue of local traps. For this reason, the MH step is often favored as a solution in discrete sampling. With the MH step and fixed step sizes, the sampler can make large jumps to facilitate global exploration. This feature is essential for navigating highly structured state spaces, where the sampler needs flexibility to move between distant states.

In practice, high-temperature samplers may have difficulty exploiting certain regions due to abrupt exploration, which requires excessive time to fully characterize local modes and achieve mixing.

\subsection{Sample Swaps between Discrete Samplers }

\begin{wrapfigure}{r}{0.35\textwidth}   
\vspace{-6.0 em}
\begin{minipage}{0.35\textwidth}
\centering
\begin{algorithm}[H]
   \caption{DREXEL and DREAM}
   \label{alg:dream}
   \footnotesize
   \textbf{Input} Step Sizes $\alpha_1$, $\alpha_2$\\
   \textbf{Input} Temperatures $\tau_1$, $\tau_2$\\
   \textbf{Input} Swap Intensity $\rho > 0$\\
   \textbf{Input} Initial Samples $\boldsymbol \theta_0^{(k)}\in \Theta$, $k=1, 2$
   \begin{algorithmic}[1]
   \State \textbf{For} $i=1,2,\cdots, I$ \textbf{do}
       \State \ \ \ \ \textcolor{darkred}{\textbf{Sampling Steps:}}
       \State \ \ \ \ \textbf{For} $k=1,2$ \textbf{do:}
       \State \ \ \ \ \ \ \ \ \textbf{For} $d=1,2,\cdots,\mathbf{d}$ \textbf{do:}
               \State\ \ \ \ \ \ \ \ \ \ \ \ Construct $q^{(k)}_{d}(\boldsymbol \theta^{(k)} \mid \boldsymbol \theta^{(k)}_{i})$ following \eqref{eq:replica_discrete}
               \State\ \ \ \ \ \ \ \ \ \ \ \ Sample $\omega^{(k)}_{d}\sim q^{(k)}_{d} ( \cdot \mid \boldsymbol \theta^{(k)}_{i})$
       \State \ \ \ \ \ \ \ \ \textbf{End For} 
       \State \ \ \ \ \textbf{End For} 
       
       \State \ \ \ \ \textcolor{darkgreen}{\textbf{MH Steps (for DREAM):}}
       \State \ \ \ \ \textbf{For} $k=1,2$ \textbf{do:}
       \State \ \ \ \ \ \ \ \ Compute ${\mathcal{A}}(\boldsymbol\theta_i^{(k)}, \boldsymbol\theta_{i+1}^{(k)})$ following \eqref{eq:mh_langevin}
       \State \ \ \ \ \ \ \ \ Generate $u\sim U[0,1]$
       \State \ \ \ \ \ \ \ \ Set $\boldsymbol{\theta}^{(k)}_{i+1}\leftarrow \omega^{(k)}$ if $u\leq {\mathcal{A}}$ else $\boldsymbol{\theta}^{(k)}_{i+1}\leftarrow \boldsymbol \theta^{(k)}_{i}$
       \State \ \ \ \ \textbf{End For} 
       
       \State\ \ \ \  \textcolor{darkblue}{\textbf{Swapping Steps:}}
       \State \ \ \ \ Generate $u\sim U[0,1]$
       \State \ \ \ \ Compute $\tilde S(\boldsymbol \theta^{(1)}_{i+1},\boldsymbol \theta^{(2)}_{i+1})$ following \eqref{eq:swap_history}
       \State \ \ \ \ Swap $\boldsymbol \theta^{(1)}_{i+1}$ and $\boldsymbol \theta^{(2)}_{i+1}$ if {$u\leq \rho\min\left\{1, \tilde S\right\}$}
   \State \textbf{End For} 
   \end{algorithmic}
   \textbf{Output} Samples $\{\boldsymbol \theta^{(1)}_i\}_{i=1}^{I}$
\end{algorithm}
\end{minipage}   
\vspace{-2.5em}
\end{wrapfigure}
A typical solution is to implement a swap function that enables sample exchanges between samplers at different temperatures. This helps cross energy barriers by combining the exploration of high-temperature samplers with the exploitation of low-temperature ones, which improves mixing rates.

The na\"ive swap function \eqref{eq:swap_naive} of reMCMC relies on energy calculations at the current samples and corresponding temperatures. However, it is not practical to handle large-scale data in mini-batch settings. Intuitively, while $\tilde U(\btheta_{i+1}^{(1)})$ and $\tilde U(\btheta_{i+1}^{(2)})$ are both unbiased in mini-batches, a non-linear transformation of these estimators fail to provide an unbiased estimator for $S(\btheta_{i+1}^{(1)}, \btheta_{i+1}^{(2)})$ \citep{deng2020non}. Under normality assumption for the energy estimate, we consider a bias correction term: 
\begin{equation}\label{eq:swap_bias_correct}
    \tilde S(\btheta_{i+1}^{(1)}, \btheta_{i+1}^{(2)}) = e^{\left( \frac{1}{\tau_2} - \frac{1}{\tau_1} \right) \left[U(\btheta_{i+1}^{(1)}) - U(\btheta_{i+1}^{(2)}) + \left( \frac{1}{\tau_1} - \frac{1}{\tau_2} \right) \sigma^2 \right]},
\end{equation}
where $\sigma^2$ compensates for noise in the stochastic gradient and removes swap bias. This adjustment ensures that the swap function behaves as a Martingale and matches the expected value obtained from exact gradients. Although this correction is not strictly necessary in discrete sampling, we retain this design in practice and examine the potential need for bias correction in the experiments. The bias-corrected versions of DREXEL and DREAM are referred to as bDREXEL and bDREAM.

When reMCMC is applied to discrete spaces, a notable challenge arises: the decaying step sizes commonly employed in continuous settings are not applicable.
% When applying reMCMC to discrete settings, we encounter another issue that decaying step sizes, which are common in continuous settings, are not applicable. 
To ensure asymptotic convergence to the target distribution with fixed step sizes, we must maintain detailed balance not only between the low-temperature and high-temperature samplers but also between the current and next output samples.
% Unlike \eqref{eq:swap_bias_correct}, which are commonly used in reMCMC, the discrete case requires maintaining a detailed balance 
% % \ruqi{may explain what local balance means. also, do you mean "detailed balance" here? not sure how this part is related to local balance} 
% when exploring the energy landscape. 
The swap designs in \eqref{eq:swap_naive} and \eqref{eq:swap_bias_correct}, however, overlook energy and temperature differences. This potentially violates detailed balance and slows down mixing in discrete sampling tasks. To mitigate the imbalance, we propose a swap function tailored for discrete sampling:
\begin{equation}\label{eq:swap_history}
    \tilde S(\btheta_{i+1}^{(1)}, \btheta_{i+1}^{(2)}\mid \btheta_{i}^{(1)}, \btheta_{i}^{(2)}) = e^{\left( \frac{1}{\tau_2} - \frac{1}{\tau_1} \right) \left[ U(\btheta_{i+1}^{(1)}) + U(\btheta_{i}^{(1)}) - U(\btheta_{i+1}^{(2)}) - U(\btheta_{i}^{(2)}) \right] }.
\end{equation}

This swap function incorporates energy estimates at the last samples, which respects the energy landscape and preserves detailed balance.
% \ruqi{can you explain why it is tailored for discrete sampling?}
Importantly, since the previous samples are treated as constants during the swap, the detailed balance between replicas remains unaffected. We will demonstrate how this correction guarantees asymptotic convergence to the target distribution in the next section.

% where $\btheta_{i}^{(1)}$ and $\btheta_{i}^{(2)}$ are the parameters from the previous iteration. This correction ensures that the sampler respects detailed balance over multiple iterations, improving long-term sampling accuracy.
\textbf{The Proposed Algorithms.} As outlined in Algorithm \ref{alg:dream}, we present DREXEL and DREAM for discrete sampling. The approaches employ two DLSs with distinct temperatures and step sizes, which allows for sample swaps between them. At each iteration, the current samples are updated, followed by MH steps in DREAM. The swap mechanism exchanges samples when the high-temperature sampler locates a lower-energy mode. After $I$ iterations, the low-temperature sampler outputs samples to characterize the energy landscape. This approach, discussed further in \ref{subsec:non_aymptotic}, improves the mixing rate over DLSs by balancing exploration and exploitation.

\section{Theoretical Analysis}\label{sec:analysis}

In the previous section, we introduced DREXEL and DREAM, which use factorization to allow parallel updates and employ swap mechanisms to improve non-convex exploration. While these features are beneficial, the overall performance heavily relies on their convergence properties and theoretical guarantees. In this section, we provide asymptotic convergence guarantees for DREXEL (i.e. the version without the MH correction).
% The analysis offers critical insights into algorithmic behavior that are not captured by empirical tests alone, which contributes to algorithm design and prevents misleading conclusions from empirical results.
% Failure to converge to the target distribution, even with MH corrections \ruqi{this reads weird. the goal of using MH correction is to ensure the chain converges to the target distribution.}, can undermine stability and introduce bias. 
% Even with MH corrections, failure to converge to the target distribution can cause the MH correction to fail, which undermines the algorithm's stability and potentially yields biased results.
% In the last section, we introduced DREXEL and DREAM, which utilize factorization to perform parallel updates and sampler swap mechanisms to enhance non-convex exploration. While these features are advantageous, the overall effectiveness depends on the algorithms' convergence properties and theoretical support. Even with MH steps, failing to converge to the target distribution would increase the sample rejection and undermine the algorithm’s reliability. 
% This section offers a formal theoretical guarantee for the proposed methods. 
% The following analysis focuses on DREXEL, as DREAM incorporates MH steps to reduce bias and improve convergence. 
% Note that DREAM further includes MH steps to offer stronger convergence guarantees, this analysis focuses on DREXEL without additional MH corrections.
% \ruqi{what does this sentence mean?}.

\subsection{Asymptotic Convergence on Log-Quadratic Distributions}\label{subsec:non_aymptotic}

Our focus is first on the asymptotic behaviors of DREXEL. The analysis aims to show that as step sizes approach zero, the algorithm exhibits zero asymptotic bias, which ensures accurate sampling from the target distribution. Specifically, we focus on log-quadratic energy $\pi(\btheta) \propto \exp{\left(\btheta^\intercal \bJ \btheta + \bm b^\intercal \btheta \right)}$, where $\bJ \in \mathbb{R}^{\mathbf{d} \times \mathbf{d}}$ is a symmetric matrix, and $\bm b \in \mathbb{R}^{\mathbf{d}}$ is a vector. If $\bJ$ is asymmetric, we apply spectral decomposition to obtain a symmetric matrix, which enables an analytically tractable solution.

\citet{zhang2022langevin} showed that DLS with temperature 1 is reversible for log-quadratic energy distributions when the step size is sufficiently small. However, this result does not directly extend to the proposed algorithm, as the swap mechanism \eqref{eq:swap_naive} introduces potential imbalances. This imbalance is due to discontinuous transitions between neighboring states in discrete space, which makes the swap acceptance rule insufficient to maintain a detailed balance. This further introduces bias during the sampling process and leads to inaccurate modeling. To address this, we carefully control the swap probability in \eqref{eq:swap_history} to regulate transitions between high- and low-temperature samplers.

\begin{tcolorbox}[frame empty, left = 0.0mm, right = 0.0mm, top = 0.0mm, bottom = 0.0mm]
\begin{theorem}\label{theorem:weak_converge}
Let $\alpha_1$ and $\alpha_2$ be the step sizes for the low- and high-temperature samplers, and let $q(\cdot|\btheta)$ be the Markov chain transition kernel. Suppose the target $\pi(\btheta)$ is log-quadratic, then:
\begin{itemize}[leftmargin=12 pt]
    \item The Markov chain induced by DREXEL is reversible with respect to an intermediate distribution $\tilde\pi$, i.e., for all $\btheta, \btheta^\prime\in\Theta\times\Theta$, $\tilde\pi(\btheta) q(\btheta^\prime|\btheta) = \tilde\pi(\btheta^\prime) q(\btheta|\btheta^\prime)$.
    \item As $\alpha_1, \alpha_2 \to 0$, the distribution $\tilde\pi$ converges weakly to the target distribution $\pi$. % \textcolor{red}{Specifically, for any bounded continuous function $f$, $\lim_{\alpha_1, \alpha_2 \to 0} \int f(\theta) \pi'(\theta) d\theta = \int f(\theta) \pi(\theta) d\theta$.}
\end{itemize}
\end{theorem}
\end{tcolorbox}
This analysis focuses on the state transition of the low-temperature sampler, as the high-temperature sampler only facilitates exploration and does not produce final samples. Intuitively, with probability $\rho \tilde{S}$, the next low-temperature sample is drawn from the high-temperature sampler, and with probability $\rho(1 - \tilde{S})$, it selects from the low-temperature sampler.
% the current low-temperature sampler transitions at a rate $\rho \tilde S$ when swapping to the next high-temperature sample and at a rate $\rho(1 - \tilde S)$ when transitioning to the next low-temperature sample. 
The transition simplifies to DLS without swaps and directly maintains the detailed balance, but the swap probability becomes essential for preserving this balance once swaps are considered in discrete sampling. Our designed swap function ensures that the overall transition dynamics remain balanced, as demonstrated in Appendix \ref{appendix:subsec:balance}.

\subsection{Non-asymptotic Convergence}

\begin{tcolorbox}[frame empty, left=0mm, right=0mm, top=0mm, bottom=0mm]
    \begin{theorem}\label{thm:convergence2}
        Let $q(\btheta \mid \btheta')$ be the transition probability of the DREXEL Markov chain, which is reversible with respect to the intermediate distribution $\tilde\pi$. Then $q$ is irreducible with eigenvalues $1 = \lambda_0 > \lambda_1 \ge \cdots \ge \lambda_{N^{2\mathbf{d}} - 1} \ge -1$. Furthermore, for all $\btheta \in \Theta \times \Theta$ and all $n \in \mathbb{N}_+$, the following bound on the total variation distance holds:
        \begin{equation}
            \| q^n(\cdot|\btheta) - \tilde\pi \|_{\mathrm{TV}} \le \frac{1}{2\sqrt{\tilde\pi(\btheta)}}  \, \lambda_*^n,
        \end{equation}
        where $\lambda_* = \max\{ \lambda_1, |\lambda_{N^{2\mathbf{d}} - 1}| \}$, and the total variation distance between two measures $\mu$ and $\nu$ is defined as
        \[
        \| \mu - \nu \|_{\mathrm{TV}} := \sup_{A \subseteq \Theta\times \Theta} | \mu(A) - \nu(A) |.
        \]
    \end{theorem}
\end{tcolorbox}

In this section, we apply a classical result from \cite{diaconis1991geometric} to estimate the total variation distance between the distribution of a general reversible Markov chain and its stationary distribution, in terms of the eigenvalues of the transition kernel.

\section{Experiments}\label{sec:experiment}

To illustrate the effectiveness of our approach, we evaluate the proposed samplers across distinct discrete sampling and generative tasks. Our approach is compared against baselines including 
% Gibbs With Gradient (GWG from \citep{grathwohl2021oops} 
% \ruqi{i thought grathwohl2021oops named their method "GWG"?}
% ), 
DLS (DULA and DMALA from \citet{zhang2022langevin}), Any-scale Balanced sampling (AB) 
% \ruqi{they called their method "AB". might be better to follow their abbreviation}
\citep{sun2023any}, and the Automatic Cyclical Sampler (ACS) \citep{pynadath2024gradient}. Baselines were reimplemented using official repositories. All experiments were conducted on a desktop featuring an AMD Ryzen Threadripper PRO 5955WX CPU, an RTX 4090 GPU, and 128 GB of DDR4 RAM. More details, such as experimental setups, hyper-parameters, and additional experimental results, can refer to Appendix \ref{appendix:sec:exp}.

\subsection{Sampling from 2D Synthetic Problems}

\begin{figure*}[!htbp]
\includegraphics[width=1.0\textwidth]{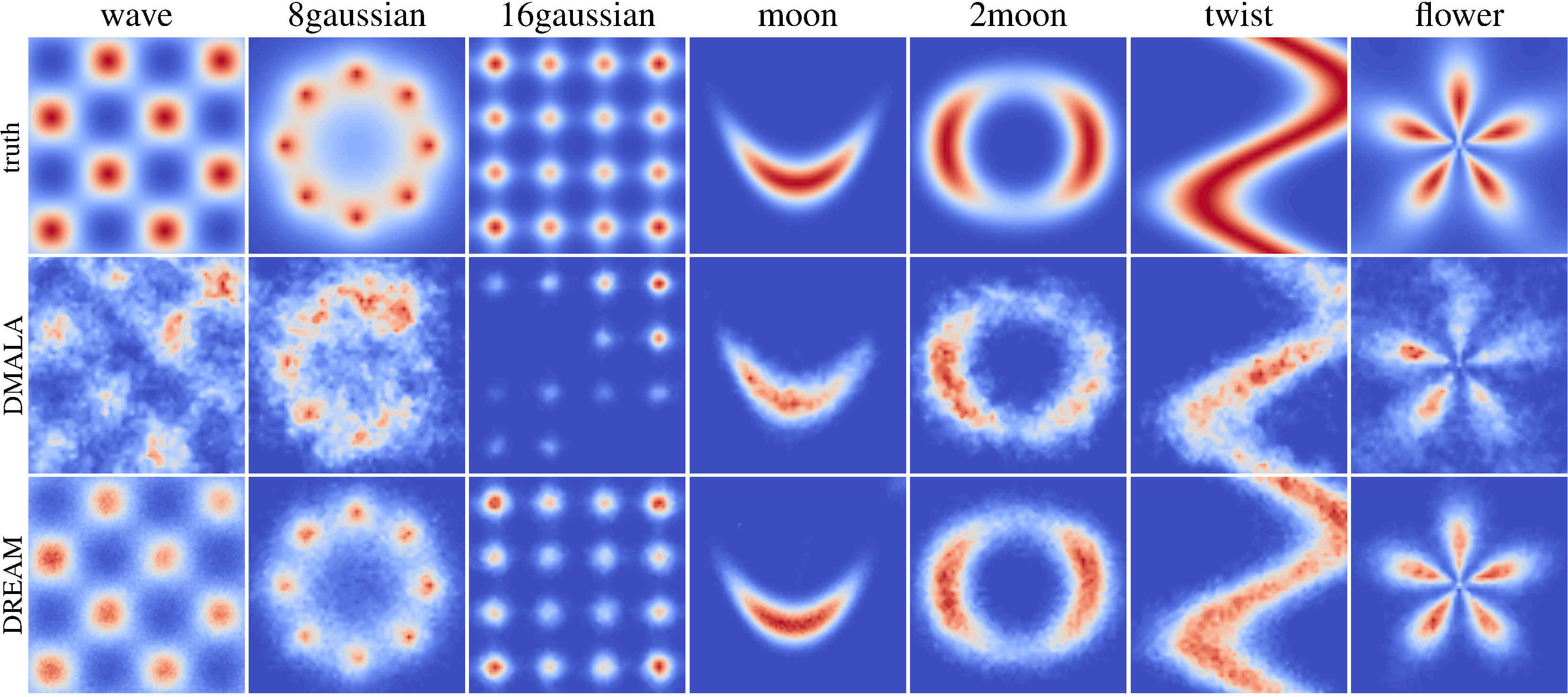}
\caption{
\textbf{Qualitative performance of discrete samplers on high-dimensional synthetic tasks.}
\textbf{Top:} Target energy landscapes (wave, 8 Gaussians, 16 Gaussians, moon, two moonx, twist, and flower) illustrate non-convex and multimodal structures with metastable regions.
\textbf{Middle and Bottom:} Empirical samples from DMALA (middle) and DREAM (right) after 100,000 iterations. DREAM consistently captures all modes across tasks, whereas DMALA fails to escape local minima and misses significant regions of the target distribution.
}\label{fig:synthetic_results}
% \vspace{-0.1 in}
\end{figure*}

We study the challenges of sampling from discrete multi-modal distributions defined over a high-dimensional domain
% $\Theta = \left\{0,1\right\}^{\mathbf{d}}$, where 
$\Theta = \left\{1,2,\dots,N\right\}^{\mathbf{d}}$, where $N=256$ and
$\mathbf{d}=51\times 51$. Each coordinate in $\Theta$ takes one of $N$ discrete values. To evaluate the energy function or its gradient, samples from $\Theta$ first need to be projected onto the discrete region between $[-2, 2] \times [-2, 2]$. Figure \ref{fig:synthetic_results} (top) highlights the challenges of approximating non-convex energy landscapes, where samplers often struggle to explore the landscapes effectively due to metastable transitions between modes or trapping in narrow energy wells—particularly under limited sample budgets. The detailed specification of the target distribution can be found in Appendix \ref{appendix:subsec_synthetic}.

\begin{table*}[]\centering
\caption{\textbf{Quantitative comparison of discrete samplers on synthetic tasks using KL divergence, MMD, and NLL.}
DREAM consistently achieves lower error across all metrics and target distributions, which indicates improved convergence and higher-quality samples compared to DMALA, AB, and ACS.}\label{tab:synthetic}
\resizebox{0.999\textwidth}{!}
{\begin{tabular}{ccccccccc}
\toprule
Metric & Sampler &   \makebox[0.08\textwidth][c]{wave} &   \makebox[0.08\textwidth][c]{8gaussian} &   \makebox[0.08\textwidth][c]{16gaussian}&   \makebox[0.08\textwidth][c]{moon} &   \makebox[0.08\textwidth][c]{2moon} &   \makebox[0.08\textwidth][c]{twist} &   \makebox[0.08\textwidth][c]{flower} \\
\midrule
\multirow{4}{*}{KL$(10^{-2})\downarrow$} 
   & DMALA & 10.891$ \pm\scriptstyle 2.408$  & 10.321$ \pm\scriptstyle 2.658$  & 8.107$ \pm\scriptstyle 3.116$   & 2.921$ \pm\scriptstyle 0.531$   & 8.544$ \pm\scriptstyle 5.227$ & 8.558$ \pm\scriptstyle 3865$ & 7.880$ \pm\scriptstyle 3.066$  \\
   & AB &  9.485$ \pm\scriptstyle 1.865$ & 3.125$ \pm\scriptstyle 1.002$ & 5.501$ \pm\scriptstyle 2.397$  & 1.570$ \pm\scriptstyle 0.434$  & 4.535$ \pm\scriptstyle 2.476$   & 2.519$ \pm\scriptstyle 0.733$ & 3.637$ \pm\scriptstyle 1.895$ \\
   & ACS   & 9.155$ \pm\scriptstyle 0.636$  & 3.357$ \pm\scriptstyle 0.219$ & 6.933$ \pm\scriptstyle 2.706$  & 1.269$ \pm\scriptstyle 0.233$  & 4.171$ \pm\scriptstyle 0.197$ & 4.031$ \pm\scriptstyle 0.264$ & \textbf{3.371}$ \pm\scriptstyle 0.066$ \\ 
   & \textbf{DREAM} & \textbf{4.975}$ \pm\scriptstyle 0.222$   & \textbf{2.393}$ \pm\scriptstyle 0.265$  & \textbf{2.509}$ \pm\scriptstyle 1.247$   & \textbf{0.500$ \pm\scriptstyle 0.119$}  & \textbf{1.067$ \pm\scriptstyle 0.154$}& \textbf{1.613$ \pm\scriptstyle 0.612$}   & 6.558$ \pm\scriptstyle 0.214$  \\ \midrule 
\multirow{4}{*}{MMD$(10^{-4})\downarrow$} 
   & DMALA & { 20.409}$ \pm\scriptstyle 1.444$ & { 13.823}$ \pm\scriptstyle 2.109$   & { 23.855}$ \pm\scriptstyle 8.301$ & 4.309$ \pm\scriptstyle 0.840$   & 6.369$ \pm\scriptstyle 2.111$ & 5.052$ \pm\scriptstyle 2.395$ & 11.053$ \pm\scriptstyle 1.604$  \\
   & AB & 13.701$ \pm\scriptstyle 3.626$   & 6.941$ \pm\scriptstyle 1.555$  & 17.201$ \pm\scriptstyle 11.780$ &  2.371$ \pm\scriptstyle 0.669$  & 4.825$ \pm\scriptstyle 1.801$ & 1.916$ \pm\scriptstyle 0.372$ & 7.247$ \pm\scriptstyle 2.036$  \\
   & ACS   &  11.818$ \pm\scriptstyle 1.216$  &  5.325$ \pm\scriptstyle 0.638$ & 11.877$ \pm\scriptstyle 1.134$   & 7.355$ \pm\scriptstyle 0.762$  & 3.638$ \pm\scriptstyle 0.390$ & 2.416$ \pm\scriptstyle 0.134$ & \textbf{6.848}$ \pm\scriptstyle 1.121$ \\
   & \textbf{DREAM} & \textbf{6.211}$ \pm\scriptstyle 0.235$   & \textbf{1.767}$ \pm\scriptstyle 0.280$  & \textbf{5.960}$ \pm\scriptstyle 0.430$   & \textbf{1.394$ \pm\scriptstyle 0.387$}  & \textbf{2.198$ \pm\scriptstyle 0.606$}   & \textbf{1.256$ \pm\scriptstyle 0.311$}   & 8.289$ \pm\scriptstyle 0.731$  \\ \midrule 
\multirow{4}{*}{NLL$(10^{-3})\downarrow$} 
   & DMALA & \textbf{7.937}$ \pm\scriptstyle 0.013$  & 5.137$ \pm\scriptstyle 0.024$ & 14.088$ \pm\scriptstyle 2.834$   & 13.181$ \pm\scriptstyle 0.010$   & 7.103$ \pm\scriptstyle 0.062$ & 7.009$ \pm\scriptstyle 0.039$ & 6.801$ \pm\scriptstyle 0.015$  \\
   & AB & 8.900$ \pm\scriptstyle 0.028$  & 5.063$ \pm\scriptstyle 0.015$ & 10.804$ \pm\scriptstyle 2.479$  & 13.160$ \pm\scriptstyle 0.020$  & 7.067$ \pm\scriptstyle 0.043$ & 6.951$ \pm\scriptstyle 0.015$ & 6.780$ \pm\scriptstyle 0.025$ \\
   & ACS   &  8.923$ \pm\scriptstyle 0.021$ & 5.063$ \pm\scriptstyle 0.008$ & 6.758$ \pm\scriptstyle 0.057$  & 13.393$ \pm\scriptstyle 0.061$  & 7.062$ \pm\scriptstyle 0.006$ & 6.970$ \pm\scriptstyle 0.007$ & 6.794$ \pm\scriptstyle 0.033$ \\ 
   & \textbf{DREAM} & {8.419}$ \pm\scriptstyle 0.010$   & \textbf{4.883}$ \pm\scriptstyle 0.003$  & \textbf{6.480}$ \pm\scriptstyle 0.025$   & \textbf{13.149}$ \pm\scriptstyle 0.006$  & \textbf{7.022$ \pm\scriptstyle 0.003$}   & \textbf{6.931$ \pm\scriptstyle 0.010$}   & \textbf{6.499}$ \pm\scriptstyle 0.004$   \\ \bottomrule
\end{tabular}}
\end{table*}

We evaluate the proposed DREAM sampler against baselines (DMALA, AB, ACS) on these tasks. All samplers are initialized by uniformly selecting a random position on the grid. DMALA uses a fixed temperature of 1.0 with step sizes between 0.020 and 0.030 (the optimal settings depend on the target distributions). AB employs $\alpha = 0.92$ and $\sigma = 0.45$, while ACS adopts 50 cycles with an initial learning rate dynamically adjusted between 0.045 and 0.060 based on the target distribution. For DREAM, the low-temperature sampler operates at 1.0 with step sizes 0.020–0.030, while the high-temperature sampler varies between 2.0 and 10.0 (adjusted per target) with larger step sizes of 0.045–0.060. With automatic differentiation for gradient computation, we generate 100,000 samples to form the empirical distributions. Each method repeats 10 times with randomized seeds for statistical robustness. Performance is measured using Kullback-Leibler (KL) divergence, Maximum Mean Discrepancy (MMD), and Negative Log-Likelihood (NLL) to quantify distributional accuracy and sample quality. Figure \ref{fig:synthetic_results} (bottom) provides a qualitative analysis, showing that DREAM can effectively capture the underlying complex distributions. In the wave, 8gaussians, and 16gaussians, DMALA (mid) captures only 50\% of modes due to its tendency to get stuck in local minima. DREAM, by contrast, recovers all modes, which reflects its robust exploration across different tasks. For quantitative evaluation, we report Kullback–Leibler (KL) divergence, maximum mean discrepancy (MMD), and negative log-likelihood (NLL) as performance metrics \citep{blessing2024beyond}. As shown in Table \ref{tab:synthetic}, DREAM consistently outperforms the baselines across all distributions.

\begin{figure}
    \centering
    
\begin{subfigure}[b]{1.0\textwidth}
\centering
\includegraphics[width=\textwidth]{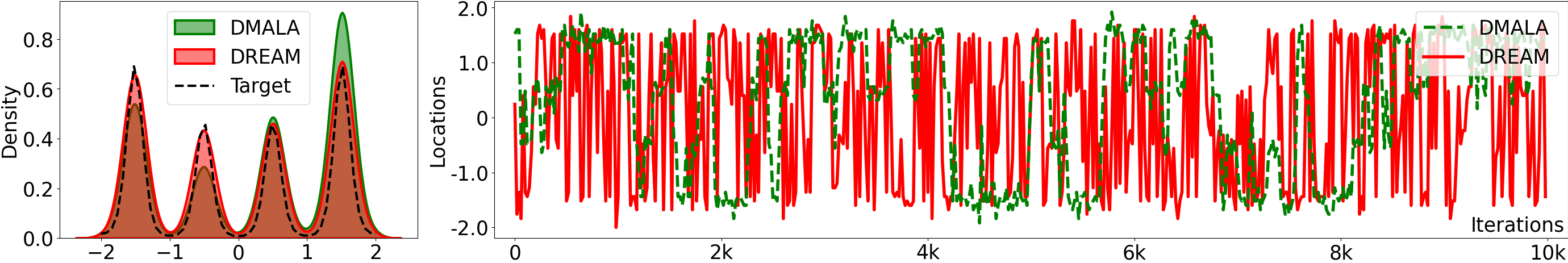}
\caption{Trace plots (100,000 samples) demonstrate that DREAM achieves significantly more non-local jumps (7.52\%) than DMALA (0.07\%), which indicates better exploration.}
\label{fig:trace}
\end{subfigure}
\vspace{0.15 in}
\begin{subfigure}[b]{0.2379154\textwidth}
\centering
\includegraphics[width=\textwidth]{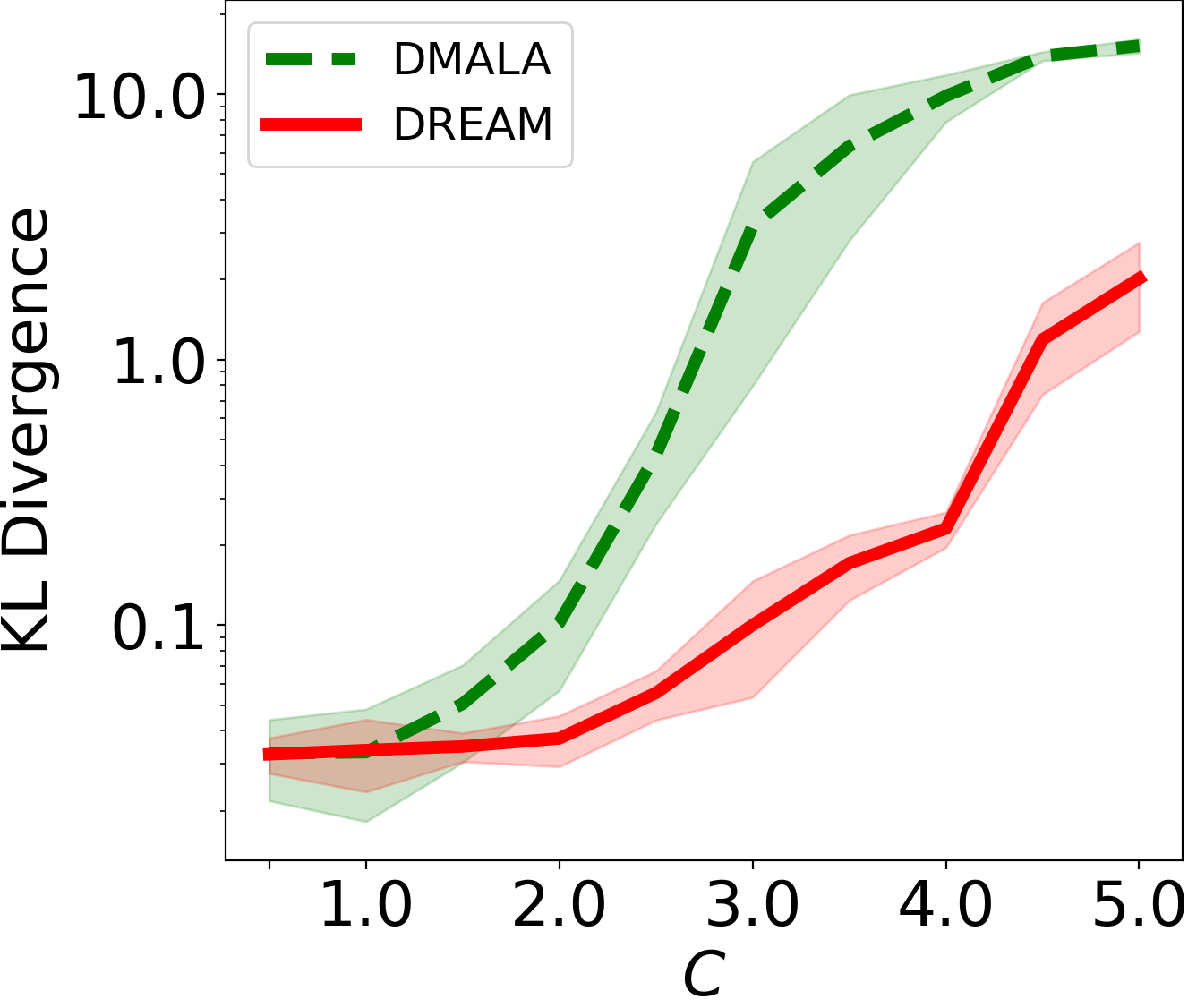}
\caption{energy barrier strength.}
\label{fig:sim_compare}
\end{subfigure}\hspace{-0.00\textwidth}
\begin{subfigure}[b]{0.2379154\textwidth}
\centering
\includegraphics[width=\textwidth]{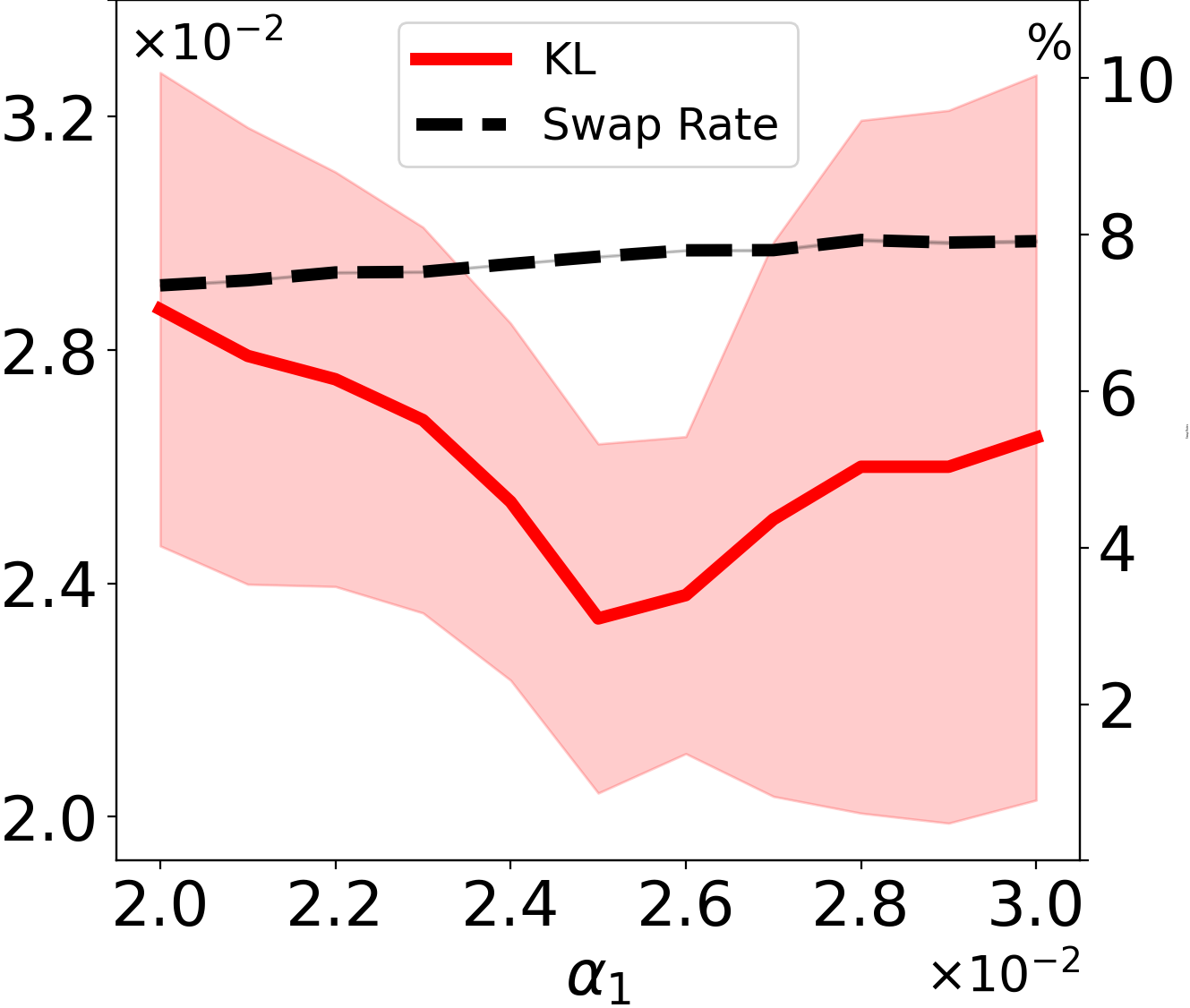}
\caption{low-temp steps.}
\label{fig:sim_low_temp_step}
\end{subfigure}\hspace{-0.00\textwidth}
\begin{subfigure}[b]{0.2379154\textwidth}
\centering
\includegraphics[width=\textwidth]{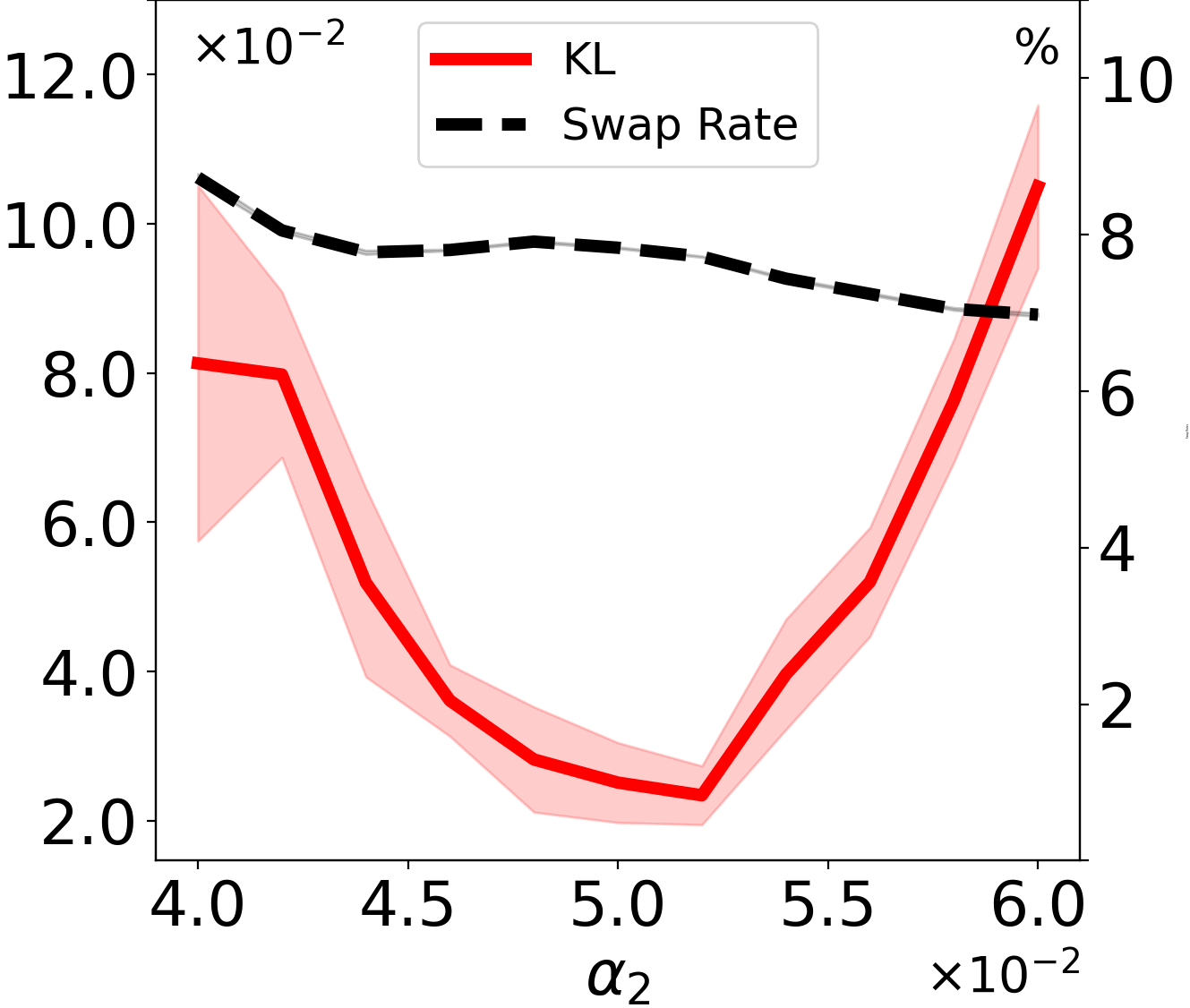}
\caption{high-temp steps.}
\label{fig:sim_high_temp_step}
\end{subfigure}\hspace{-0.00\textwidth}
\begin{subfigure}[b]{0.2379154\textwidth}
\centering
\includegraphics[width=\textwidth]{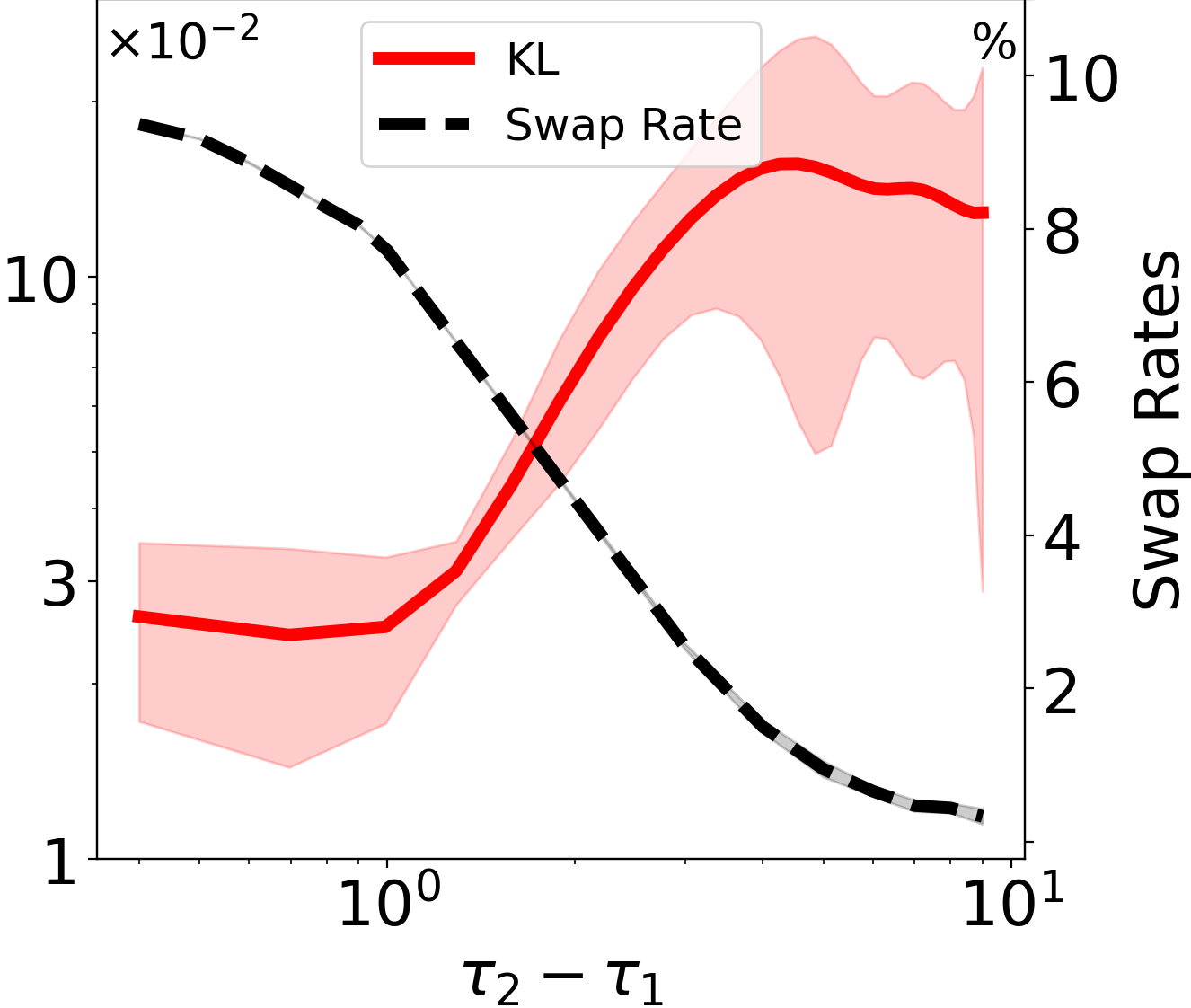}
\caption{{temperatures.}}
\label{fig:sim_high_temp}
\end{subfigure}

    \caption{\textbf{Comparative analysis of DMALA and DREAM on the 16-Gaussian task.} \textbf{(a)} Trace plots over 100,000 samples. \textbf{(b)–(e)} Ablation studies: \textbf{(b)} KL divergence across increasing energy barrier strength $C$; \textbf{(c)} sensitivity to low-temperature step size; \textbf{(d)} high-temperature step size; \textbf{(e)} high-temperature chain temperature. Results highlight that DREAM is robust across settings but sensitive to high-temperature parameters affecting swap rates.}
\end{figure}

To investigate why DREAM outperforms DMALA, we analyze their behavior on the 16 Gaussian task using empirical density and trace plots (Fig. \ref{fig:trace}). For this task, we set the low-temperature sampler to operate at 1.0 with step size 0.023 and the high-temperature sampler at 2.0 with step size 0.053. The target distribution of the 16 Gaussian can be expressed as 
\begin{equation}\label{eq:target_16gauss}
U(x, y) = \frac{x^2 + y^2}{5} - C\left[\cos(2\pi x) + \cos(2\pi y)\right],
\end{equation}
where we set $C=2.0$ here. The empirical density along the y-axis (left) illustrates that DREAM (red) aligns more closely with the target distribution (grey dashed line), while DMALA (green) exhibits significant deviations. Trace plots, plotted every 200 samples, reveal that DREAM (red solid line) transitions frequently between modes due to its high-temperature exploration and replica exchange mechanism, whereas DMALA (green dashed line) remains confined to local regions for prolonged intervals. Quantitatively, we measure non-local jumps (transitions with L2 distance greater than 1.0 within the $[-2,2]\times[-2,2]$ domain) made by DREAM and DMALA. Over 100,000 samples, DREAM achieves non-local jumps rate of 7.52\%, while DMALA attains only 0.07\%.  This stark contrast underscores DREAM's ability to navigate non-convex energy landscapes, which enables effective exploration of non-convex energy landscapes through its dual-temperature design.

We subsequently assess the robustness of DMALA and DREAM under escalating energy barriers by varying the parameter $C$ from 0.5 to 5.0 in the energy function \eqref{eq:target_16gauss}, which raises energy barriers and makes it harder for samplers to escape local minima. DMALA operates at a fixed temperature of 1.0 and step size of 0.025, while DREAM employs identical settings for its low-temperature chain and a high-temperature chain with temperature 2.0 and step size 0.053. Experiments are repeated 10 times with randomized seeds, and KL divergence metrics (mean and std) are reported (Figure \ref{fig:sim_compare}). As $C$ increases, DREAM consistently outperforms DMALA in KL, which demonstrates its ability to avoid local traps in high-barrier regimes. To investigate the sensitivity of DREAM to hyperparameters, we set $C=2.0$, and vary low-temperature step size (0.020–0.030), high-temperature step size (0.040–0.060), and high-temperature chain temperature (1.01–10.00) (Figures \ref{fig:sim_low_temp_step}–\ref{fig:sim_high_temp}). The results reveal that DREAM is sensitive to both step sizes and temperatures in the high-temperature chain. Notably, increasing the high-temperature chain temperature from 2.0 to 10.0 leads to a significant drop in the swap rate (from 10.0\% to 1.2\%, Figure \ref{fig:sim_high_temp}), which indicates that excessively high temperatures can hinder effective chain swaps. This suggests that an appropriate balance in temperature and step size is crucial for maintaining efficient exploration through chain swaps in DREAM. 

\subsection{Sampling from Ising Models}

The Ising model \citep{newman1999monte} describes a mathematical structure used to describe systems of interacting binary variables, which are commonly represented as spins in physical systems \citep{mackay2003information}. Each spin can assume binary states and interact with adjacent spins within a lattice. The interactions between these neighboring spins are governed by the energy function 
\begin{equation*}\label{eq:ising_energy}
    U(\btheta) = w\btheta^\intercal \bJ\btheta + \bm{b}^\intercal \btheta,
\end{equation*}
where $\btheta\in\left\{-1, 1\right\}^{\mathbf d}$ is binary random variable, $\bJ\in \{0, 1\}^{\mathbf d\times \mathbf d}$ is a binary adjacency matrix, $w\in\mathbb R^+$ denote the connectivity strength, and $\bm{b}\in \left\{0, 1\right\}^{\mathbf d}$ is the bias vector. 

% In Figure \ref{fig:ising_results} (left), We first show that DREAM can maintain a high acceptance rate 

% \begin{figure*}[hbtp]
% \centering
% \includegraphics[width=1.25 in]{figs/ising/log_rmse.pdf}\label{fig:ising:log_rmse:iter}
% \includegraphics[width=1.25 in]{figs/ising/log_rmse_time.pdf}\label{fig:ising:log_rmse:time}
% % \centering\includegraphics[width=1.25 in]{figs/ising/ising_swap.png}\label{fig:ising_swap}

% % \centering\includegraphics[width=1.25 in]{figs/ising/ising_logrmse.png}\label{fig:ising_logrmse}

% % \centering\includegraphics[width=1.25 in]{figs/ising/ising_step.png}\label{fig:ising_runtime}
% \caption{Ising model sampling results.}\label{fig:ising_results}
% \end{figure*}

% \centering
\begin{figure*}[!htbp]
\begin{minipage}{0.19\textwidth}
    \includegraphics[height=1.3 in]
    % {figs/ising/log_rmse_time.png}\label{fig:ising:log_rmse:iter}
    {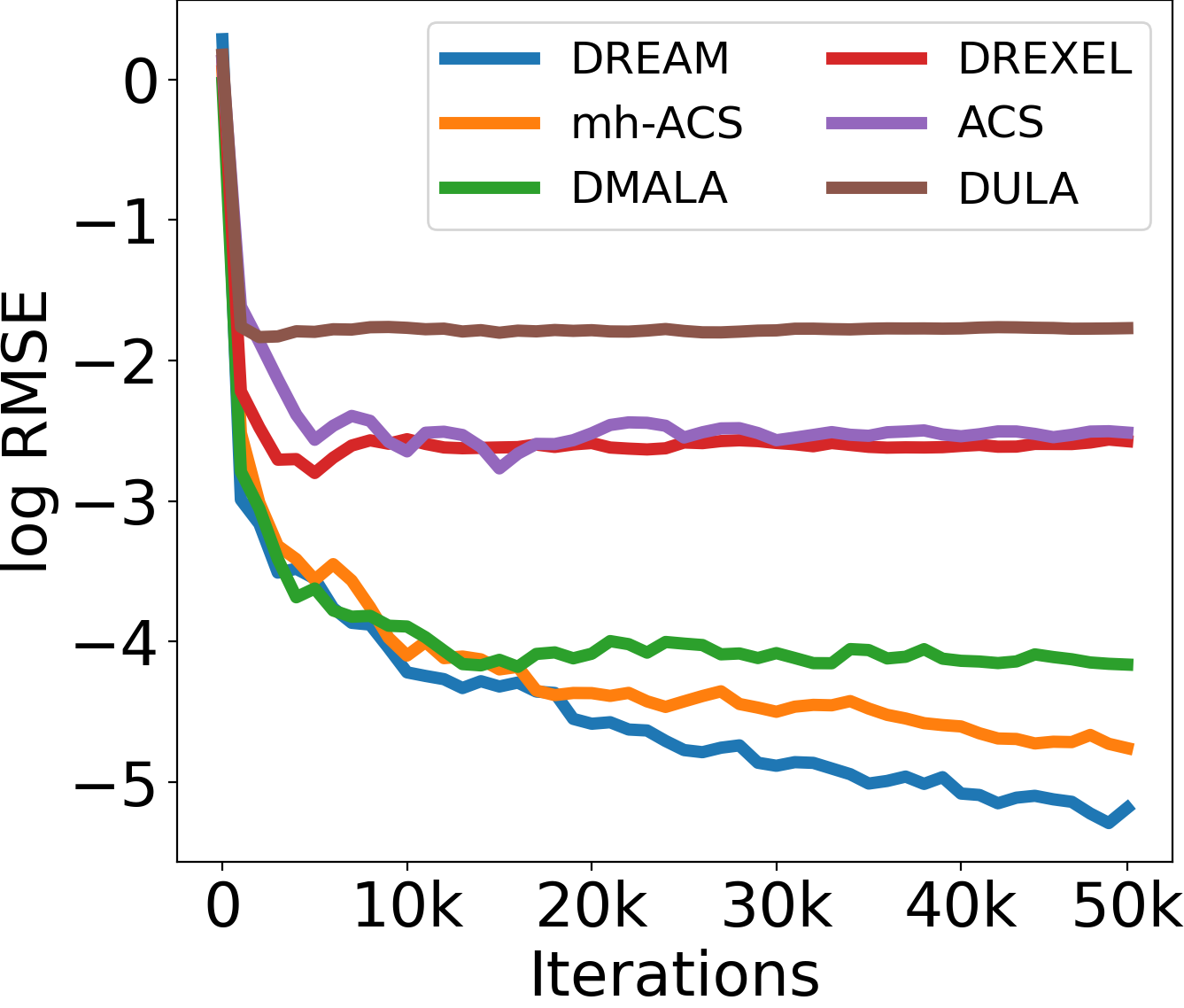}\label{fig:ising:log_rmse:iter}
\end{minipage}\hspace{0.05\textwidth}
\begin{minipage}{0.19\textwidth}
    \includegraphics[height=1.3 in]{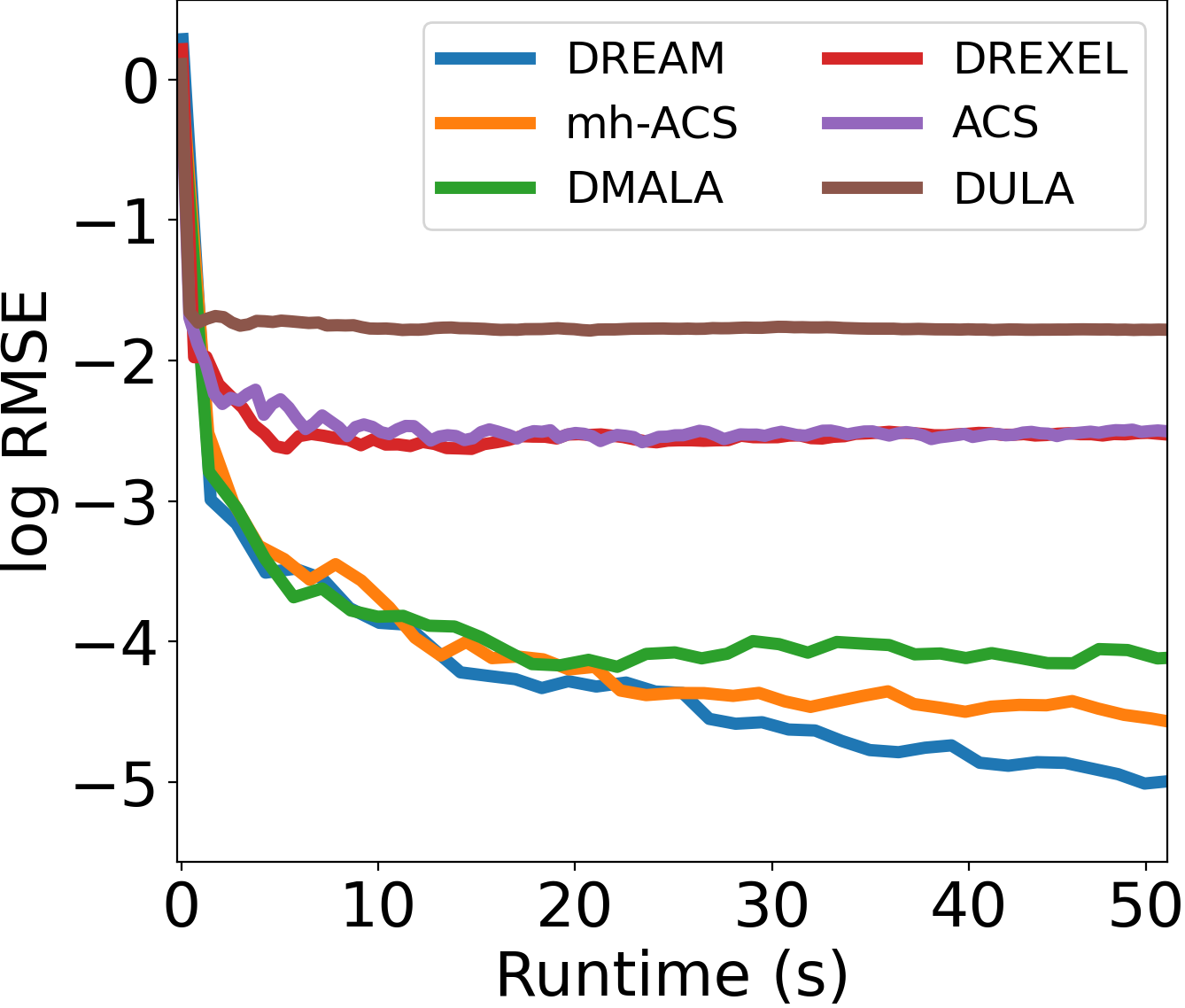}\label{fig:ising:log_rmse:time}
\end{minipage}\hspace{0.05\textwidth}
\begin{minipage}{.3\linewidth}
\resizebox{0.95\textwidth}{!}{
    \begin{tabular}{@{}lcccccc}
    \toprule
Sampler & MH & log RMSE $\downarrow$ \\ \midrule
DULA    & $\usym{1F5F6}$ & -1.769   $\pm$ 0.022\\
ACS     & $\usym{1F5F6}$ & -2.464   $\pm$ 0.084    \\
\textbf{DREXEL}  & $\usym{1F5F6}$ & \textbf{-2.544   $\bm\pm$ 0.033} \\ \midrule
DMALA   & $\usym{1F5F8}$ & -4.595   $\pm$ 0.133    \\
mh-ACS  & $\usym{1F5F8}$ & -4.691  $\pm$ 0.210    \\
\textbf{DREAM}   & $\usym{1F5F8}$ & \textbf{-4.884   $\bm\pm$ 0.204}    \\ \bottomrule
    \end{tabular}}\vspace{0.10 in}
\end{minipage}\hspace{-0.01\textwidth}
\begin{minipage}{0.19\textwidth}
    \includegraphics[height=1.3 in]{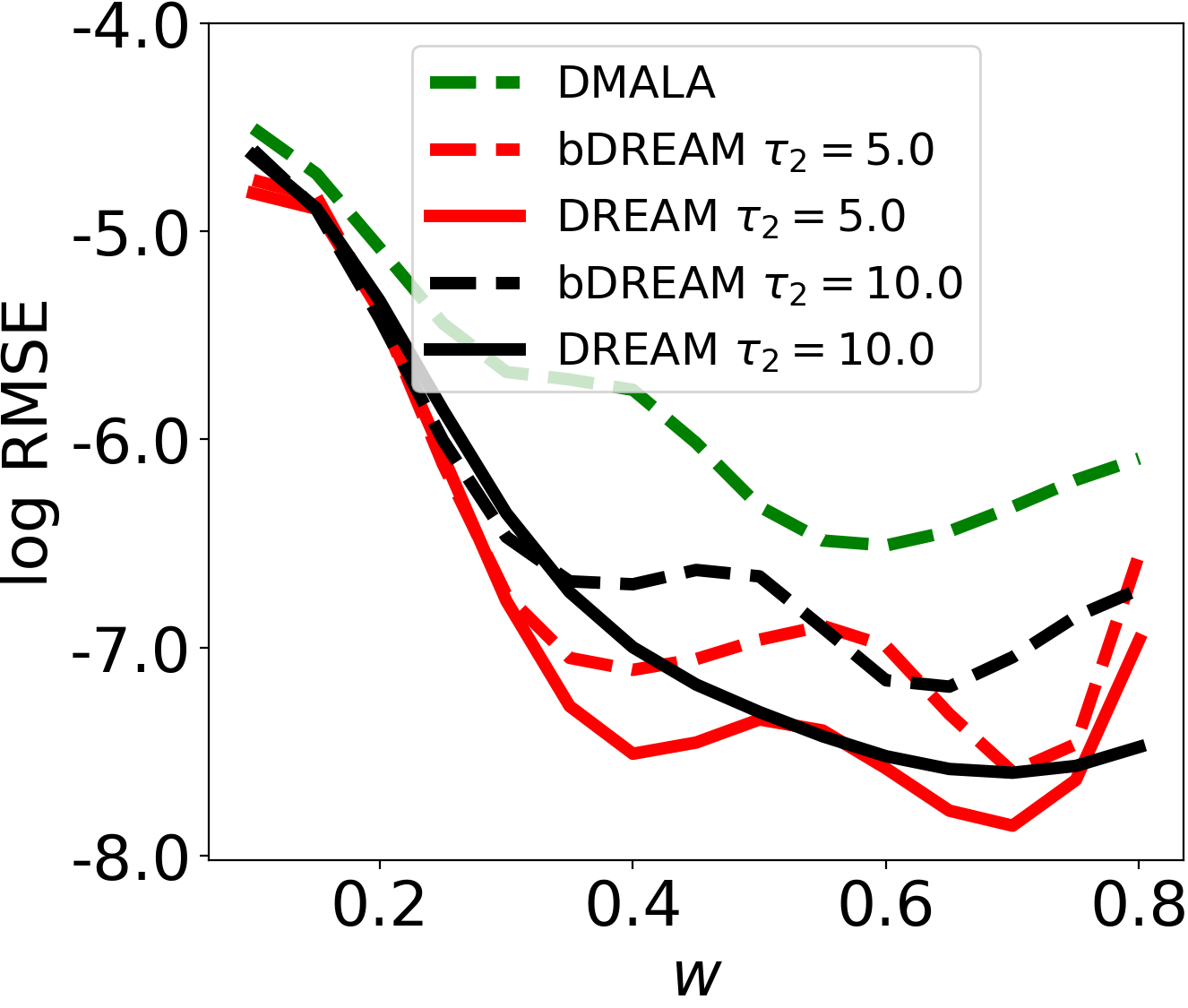}\label{fig:ising:ablation}
\end{minipage}
\vspace{-0.1 in}
\caption{\textbf{Performance of discrete samplers on 2D Ising models.} \textbf{Left:} RMSE (log scale) vs. iteration and runtime show DREXEL and DREAM significantly outperform DULA and DMALA, respectively, both with and without Metropolis–Hastings (MH) corrections.
\textbf{Middle:} Summary table reports mean ± std log RMSE over 10 runs, showing consistent performance gains from replica exchange.
\textbf{Right:} Ablation across connectivity $w \in [0.1, 0.8]$ shows that DREAM achieves peak gains near critical coupling, validating its improved exploration in strongly correlated, high-barrier regimes.
% Ising model sampling results, evaluated by log RMSE. DREAM yields the best scores.
}\label{fig:ising:results}\vspace{-0.10 in}
\end{figure*}

We first compare DREAM with other baselines to see the ability to approximate Ising models. All samplers consider a default temperature of 1.0 unless otherwise specified. DULA employed a step size of 0.20, while DMALA used 0.40. ACS implemented 10 cycles with an initial step size of 0.30, and its MH-corrected variant started at 1.0. DREXEL with small/large step sizes (0.20/0.40) and temperatures (1.0/5.0), and DREAM with steps (0.40/0.90) under identical temperature settings. Initial spin configurations are drawn from a Bernoulli distribution, where the logits are set to $P(\btheta_i = +1) \approx 0.6$ and $P(\btheta_i = -1) \approx 0.4$. For model evaluations, we set connectivity strength $ w = 0.15 $. Performance is quantified using log Root Mean Square Error (log RMSE) across 50,000 sample sizes and computational budgets ($ t =50.0 $ seconds), with 10 random seeds for statistical significance.

In Figure \ref{fig:ising:results}, DREXEL and DREAM demonstrate significant improvements over baseline methods. Without MH corrections, DREXEL reduces RMSE by over 50\% compared to DULA; With MH corrections, DREAM outperforms DMALA with a nearly 25\% reduction in RMSE. Both improvements are consistent across trials, as evidenced by low standard deviations. Furthermore, it implies that single-chain discrete samplers do not effectively exploit local modes, and the proposed samplers generally offer better and more reliable mixing rates.

We continue to investigate how connectivity strength $ w $ in Ising models affects sampler performance, specifically testing our theoretical analysis from Section \ref{sec:analysis} where we established convergence guarantees for log-quadratic energy functions. The Ising model's energy \eqref{eq:ising_energy} provides an ideal testbed as it explicitly satisfies the log-quadratic assumption. By varying $ w \in [0.1,0.8] $, we systematically increase energy barriers between spin configurations: stronger connectivity amplifies alignment penalties between adjacent spins, creating deeper local minima separated by wider activation gaps.

Our experiment compares DMALA against DREAM variants (with/without swap correction) to validate two theoretical claims: (1) DREAM improves non-convex exploration compared to DMALA; (2) the swap mechanism in \eqref{eq:swap_history} preserves detailed balance and improve mixing rates in log-quadratic systems. Each configuration undergoes 50,000 sampling iterations with tuned step sizes, repeated 10 times to measure average log RMSE. We include bDREAM to assess whether the theoretical correction benefits practical performance on exactly solvable systems.

From Figure \ref{fig:ising:results} (right), the performance gap between DREAM and baseline methods (DMALA, bDREAM) varies significantly acrosss the connectivity strength $w$ in the Ising model. In weakly coupled regimes ($w \in [0.1, 0.3]$), where energy barriers scale linearly and thermal fluctuations dominate, DREAM achieves modest improvements over DMALA, as baseline samplers retain sufficient mobility in shallow energy landscapes. Near critical coupling ($w \in [0.4, 0.7]$), where correlation lengths diverge and ferromagnetic ordering emerges, DREAM’s replica exchange mechanism becomes critical, reducing log RMSE by around $30\%$ by leveraging high-temperature chain exploration of metastable states. This peak performance aligns with Onsager’s critical temperature prediction for 2D Ising models \citep{onsager1944crystal}. Beyond criticality ($w > 0.7$), superlinear barrier growth degrades all methods, though DREAM maintains relative superiority through enhanced state swaps between frozen clusters. Notably, increasing $\tau$ from $5.0$ to $10.0$ diminishes performance (e.g., $\tau=10.0$: $-7.48$ vs $\tau=5.0$: $-7.86$ for DREAM at $w=0.7$), which highlights the sensitivity of swap efficiency to temperature scaling. These results empirically demonstrate that replica exchange optimally balances exploration and exploitation near phase transitions, where conventional samplers stagnate due to diverging energy barriers.

\subsection{Sampling from Restricted Boltzmann Machines}

Restricted Boltzmann Machines (RBMs) are generative stochastic neural networks designed to model complex distributions over discrete data \citep{fischer2012introduction}. RBMs typically consist of binary-valued hidden and visible units, where the visible units represent observed data and the hidden units capture latent dependencies in the data. The energy function 
% \begin{equation*}\label{eq:rbm_energy}
    $U ( \btheta) = \log\left[1+\exp\left( \bJ \btheta + \bm c\right)\right] + \bm b^{\top} \btheta,$
% \end{equation*}
where $\btheta \in \{0, 1\}^{\mathbf d}$ represents the binary state vector for the visible layer, $\bJ \in \mathbb{R}^{\mathbf m\times \mathbf d}$ is the weight matrix, $\bm c \in \mathbb{R}^{\mathbf m}$ and $\bm b \in \mathbb{R}^{\mathbf d}$ denote biases for hidden units and visible units correspondingly.

\begin{figure*}[!htbp]\centering
\begin{minipage}{3.5cm}\vspace{0.10 in}
    \includegraphics[width=1.8 in]{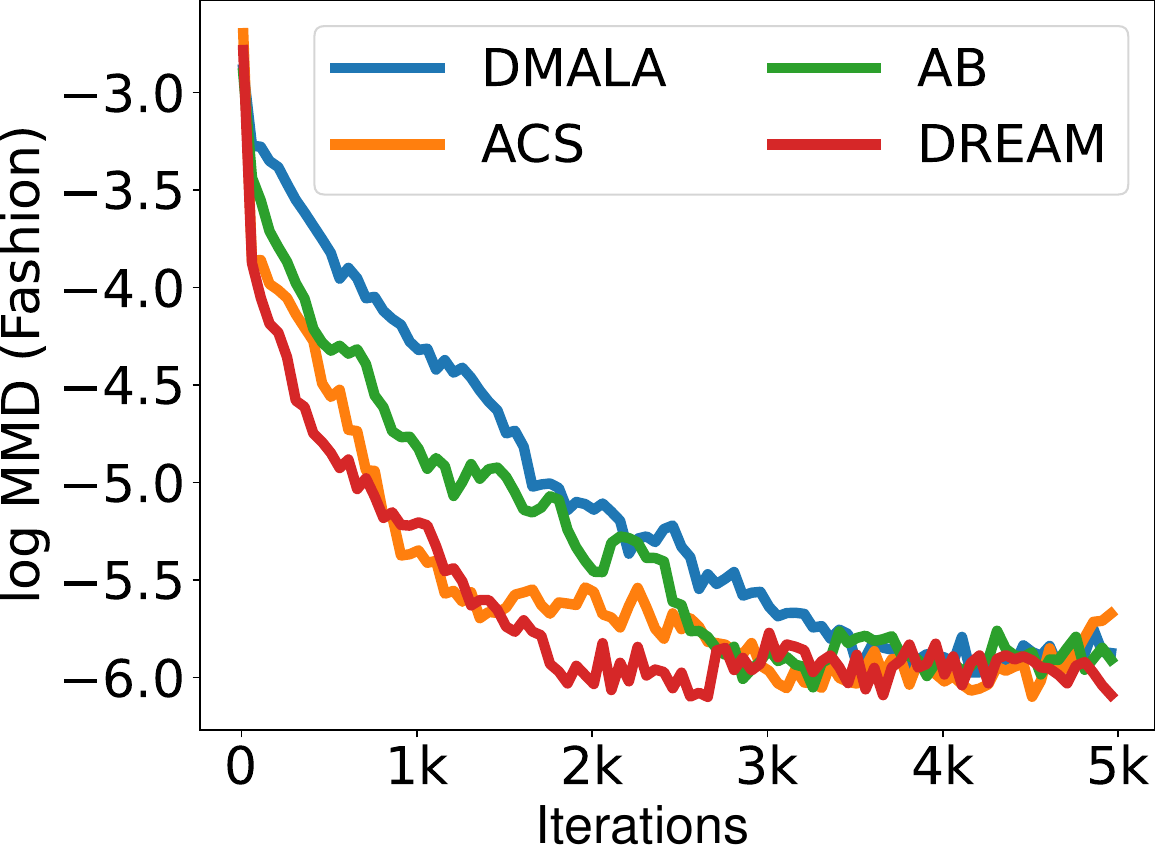}\label{fig:rbm:log_mmd}
\label{fig:fig2}  
\end{minipage}$\quad\quad\quad$
\begin{minipage}{.6\linewidth}\vspace{0.10 in}
\resizebox{0.99\textwidth}{!}{
\begin{tabular}{lcccccc} \toprule
Dataset     & \makebox[0.12\textwidth][c]{DMALA}   & \makebox[0.12\textwidth][c]{ACS} & \makebox[0.12\textwidth][c]{AB} & \makebox[0.12\textwidth][c]{\textbf{DREAM}} \\ \midrule
MNIST   & -6.248$\pm\scriptstyle 0.059$  & -6.325$\pm\scriptstyle 0.039$  & -6.328$\pm\scriptstyle 0.060$  & \textbf{-6.349}$\pm\scriptstyle {0.061}$   \\
eMNIST  & -6.082$\pm\scriptstyle 0.039$  & -6.118$\pm\scriptstyle 0.071$  & -6.119$\pm\scriptstyle 0.048$  & \textbf{-6.148}$\pm\scriptstyle {0.077}$    \\
kMNIST  & -5.821$\pm\scriptstyle 0.038$ & -5.836$\pm\scriptstyle 0.061$ & -5.841$\pm\scriptstyle 0.065$ & \textbf{-5.875}$\pm\scriptstyle {0.040}$   \\
Fashion & -5.808$\pm\scriptstyle 0.177$  & -5.835$\pm\scriptstyle 0.118$  & -5.869$\pm\scriptstyle 0.145$  & \textbf{-5.901}$\pm\scriptstyle {0.077}$    \\
Omniglot & -6.536$\pm\scriptstyle 0.099$  & -6.565$\pm\scriptstyle 0.078$  & -6.560$\pm\scriptstyle 0.074$ & \textbf{-6.626}$\pm\scriptstyle {0.057}$    \\ 
Caltech & -5.740$\pm\scriptstyle 0.078$  & -5.765$\pm\scriptstyle 0.045$  & -5.771$\pm\scriptstyle 0.047$  & \textbf{-5.810}$\pm\scriptstyle {0.085}$    \\ \bottomrule 
\end{tabular}}
\vspace{0.25 in}
% \captionof{table}{Some table}
\label{tab:tab}  
\end{minipage}
\vspace{-0.30 in}
\caption{\textbf{RBM sampling performance across six datasets, evaluated by log MMD.}
\textbf{Left:} Log MMD comparisons show DREAM achieves the lowest values across all datasets, indicating superior convergence to the RBM equilibrium distribution.
\textbf{Right:} Summary table reports mean $\pm$ std of log MMD over 10 runs. DREAM consistently outperforms other methods, achieving relative MMD reductions of 8.2–12.3\% over DMALA, with low variance confirming robust performance.
% RBM sampling results, quantified by log MMD. DREAM outperforms across various datasets.
}\label{fig:rbm:results}\vspace{-0.1 in}
\end{figure*}

We trained RBMs with 500 hidden units on six datasets: MNIST, eMNIST, kMNIST, Fashion, Omniglot, and Caltech Silhouettes datasets using contrastive divergence (CD) \citep{hinton2002training}. For each dataset, we first tuned two critical hyperparameters: the learning rate and the number of Gibbs sampling steps in CD. We evaluated pairs of these hyperparameters across 10 repeated trials under fixed DMALA sampler settings, which selects the configuration that achieved the lowest MMDs between model samples and training data.

Once the optimal learning rates and CDs were determined, we benchmarked four samplers: DMALA, ACS, AB, and DREAM. All experiments used a default temperature of 1.0. For DMALA, step sizes were explored in the range [0.15, 0.25]. ACS employed 20 cycles with initial step sizes between 1.00 and 1.60. For AB, the weights $\alpha$ range from 0.60 to 1.00, and the scales of the proposal distributions $\sigma$ are tuned between 0.40 and 0.80. DREAM utilizes high temperatures in [4.8, 8.0], low-temperature steps in [0.15, 0.25], and high-temperature steps in [0.30, 0.45]. The detailed configuration of DREAM can be found in Table \ref{tab:rbm_parameters} in the Appendix. The sample distribution is initialized using a Bernoulli distribution with equal probabilities for all visible units. To evaluate sampler performance, we repeat the test 10 times with different random seeds and compute MMD values (on average) between samples generated by each method and those from a structured Block-Gibbs sampler tailored to RBMs.

Figure \ref{fig:rbm:results} demonstrates that DREAM consistently achieves the lowest MMD values across all datasets, which indicates superior convergence in sampling from RBMs. To quantify improvements over DMALA, we compute relative reductions in MMD on the original scale. For MNIST, DREAM reduces MMD by 9.7\% compared to DMALA.
% (from $e^{-6.248}\approx 1.95\cdot 10^{-3}$ to $e^{-6.349}\approx 1.76\cdot 10^{-3}$)
Similar improvements are observed across datasets, ranging from 8.2\% (eMNIST) to 12.3\% (Omniglot).
% , with small standard deviations confirming statistical consistency
AB and ACS perform competitively but exhibit marginally higher MMD values than DREAM, which suggests slightly less stable sampling. DMALA trails behind all methods, which highlights the limitations of fixed-step sampling in high-dimensional spaces. These results underscore DREAM’s robustness and efficiency in approximating the RBM equilibrium distribution.

\subsection{Learning Energy-Based Models}
% \ruqi{why do the names change to "bDREXEL", "DREXEL"? are the algorithms used here different from those in previous tasks?}

Deep Energy-Based Models (EBMs) \citep{ngiam2011learning, bond2021deep} are a class of probabilistic models where the energy function is parameterized by a ResNet \citep{he2016deep}. Specifically, the probability of a data point $\bm x$ is given by $P_\btheta(\bm x) = \exp\left[E_\btheta(\bm x)\right] / Z_\btheta$, where $E_\btheta(\bm x)$ is the energy function parameterized by $\btheta$, and $Z_\btheta = \mathbb E_{\btheta\sim\Theta} \exp\left[E_\btheta(\bm x)\right]$ normalizes the distribution.

With DULA and DMALA as baselines, we evaluate DREXEL and DREAM\footnote{Unless stated otherwise, DREXEL and DREAM refer to samplers based on the swap design from \eqref{eq:swap_history}. bDREXEL and bDREAM apply the swap function described in \eqref{eq:swap_bias_correct}.} by learning Deep EBMs. Initial samples are drawn from a Bernoulli distribution parameterized by the empirical mean of the first batch in the dataset. During training, we consider ResNet-64 backbone, which is optimized with Adam (learning rate = 0.001) for 20,000 iterations, employing a batch size of 256. The intractable likelihood gradient of the model is approximated through Persistent Contrastive Divergence \citep{tieleman2009using}, while a replay buffer \citep{du2019implicit} containing 1,000 past samples is implemented to improve both the efficiency and stability of the training process. Each sampler runs for 40 steps per iteration. The baseline samplers (DULA and DMALA) were configured with fixed step sizes (0.08 and 0.10, respectively) and a temperature of 1.0. For replica exchange methods, DREXEL and its bias-corrected variant (bDREXEL) utilized a low-temperature sampler with step size 0.05 and temperature 1.0, paired with a high-temperature sampler (step sizes 0.14–0.16, temperature 5.0). Similarly, DREAM and bDREAM employed a low-temperature chain with step sizes 0.08–0.11 and temperature 1.0, coupled with a high-temperature chain (step sizes 0.15–0.30, temperature 5.0). Upon completing training, Annealed Importance Sampling \citep{neal2001annealed} is conducted with DULA to estimate the test log-likelihoods and ensure robust convergence diagnostics.

\begin{table*}[ht]
\centering
\caption{
\textbf{Test log-likelihoods of Deep EBMs trained on binary image datasets.} DREAM achieves the best performance across all datasets, with particularly large gains on Omniglot and Caltech Silhouettes. These results highlight the effectiveness of the MH corrections and DREAM in improving sample quality and convergence for discrete energy-based models. Lower log-likelihood indicates better performance.
}
\label{tab:ebm:20k}
% \vspace{-0.1 in}
\resizebox{0.999\textwidth}{!}{
\begin{tabular}{lcccccc} \toprule
Dataset     &  DULA   & DMALA   & bDREXEL & bDREAM & DREXEL & \textbf{DREAM} \\ \midrule
Static MNIST        & $-84.641\scriptstyle \pm 0.407$  & $-85.152\scriptstyle \pm 0.547$  & $-85.686\scriptstyle \pm 0.802$  & $-84.892\scriptstyle \pm 1.267$  & $-84.047\scriptstyle \pm 0.312$   & $\textbf{-83.711}\scriptstyle \pm \textbf{0.930}$    \\
Dynamic MNIST       & $-86.634\scriptstyle \pm 0.525$  & $-84.799\scriptstyle \pm 0.598$  & $-86.977 \scriptstyle \pm 0.641$  & $-85.158 \scriptstyle \pm 1.065$  & $-84.043 \scriptstyle \pm 0.378$   & $\textbf{-83.003} \scriptstyle \pm \textbf{0.355}$   \\
Omniglot    & $-118.627 \scriptstyle \pm 5.402$ & $-111.860 \scriptstyle \pm 1.155$ & $-102.489 \scriptstyle \pm 2.736$ & $-100.117 \scriptstyle \pm 2.167$ & $-101.973 \scriptstyle \pm 1.897$  & $\textbf{-98.525} \scriptstyle \pm \textbf{1.782}$  \\
Caltech   & $-108.708 \scriptstyle \pm 4.325$  & $-107.895 \scriptstyle \pm 5.497$ & $-108.210 \scriptstyle \pm 7.722$  & $-107.998 \scriptstyle \pm 11.856$  & $-93.543 \scriptstyle \pm 2.795$   & $\textbf{-92.004} \scriptstyle \pm \textbf{2.600}$   \\ \bottomrule 
\end{tabular}}
\end{table*}

We trained Deep EBMs on binary images from Static MNIST, Dynamic MNIST, Omniglot, and Caltech Silhouettes datasets. The test log-likelihoods for trained models across different samplers are recorded in Table \ref{tab:ebm:20k}. 

Among the samplers, DREAM consistently achieved the highest log-likelihoods across all datasets, with notable improvements on Omniglot and Caltech Silhouettes, which outperforms baselines by 11.9–14.7\%. For MNIST variants, DREXEL and DREAM showed competitive performance, particularly on Static MNIST, though DREAM maintained a clear edge in balancing exploration and exploitation. In contrast, bDREXEL and bDREAM generally performed worse across datasets, with DREXEL and DREAM demonstrating superior empirical performance. These findings confirm that MH corrections are essential for improving sampling fidelity in discrete tasks, as uncorrected replicas exhibited slower convergence. Also, the proposed swap mechanism in \eqref{eq:swap_history} is effective at correcting imbalance and yielding better log-likelihood estimates across diverse image datasets. This is particularly evident in datasets like Omniglot, where DREAM reduced the NLL gap to DMALA by 13.3\%.

\section{Conclusion and Discussion}\label{sec:conclude}

In this work, we addressed the challenge of balancing global exploration and local exploitation in non-convex discrete energy landscapes by proposing DREXEL and DREAM. These samplers integrate DLS with replica exchange to overcome the limitations of traditional samplers, which tend to get trapped in local modes due to reliance on local gradients and small disturbances. 

The proposed samplers are theoretically shown to satisfy detailed balance, which ensures correctness with respect to the target distribution. We further establish both asymptotic and non-asymptotic convergence guarantees. The empirical evidence suggests that the proposed samplers and swap mechanism significantly improve exploration and mixing in non-convex discrete spaces. Furthermore, while DREXEL maintains detailed balance throughout the process, MH corrections are critical for optimizing performance in certain tasks.

Our current work focuses on designing a single low-temperature and a single high-temperature sampler. Future research could extend this framework by introducing multiple parallel samplers to enhance exploration. We will also study theoretical guarantees to quantify the acceleration effect of the swap mechanism in the future.

\bibliography{main}
\bibliographystyle{tmlr}

\appendix

\section{Futher Discussion on Related Work}

\textbf{Related Discrete Methods} have been made in modeling and optimizing discrete distributions. \citet{zhang2022generative} introduced energy-based generative flow networks to amortize expensive MCMC exploration into a fixed number of actions. Discrete Diffusion Models \citep{sun2022score, loudiscrete} extended continuous-time diffusion models to discrete spaces with well-defined score functions for discrete variables. \citet{wen2024hard} proposed an efficient gradient-based discrete optimization method for generative models. These approaches to discretizing continuous methods and handling complex discrete data offer valuable insights for developing DLSs.

\textbf{Stochastic Gradient Langevin MCMC} \citep{welling2011bayesian} has become a favored MCMC method in big data due to its effective transition from optimization to sampling. However, its lack of adaptive step sizes to the energy curvature limits the use of this crucial information. To further leverage curvature information, Quasi-Newton methods \citep{ahn2012bayesian, simsekli2016stochastic} exploit curvature information by adjusting step sizes, while Hamiltonian Monte Carlo \citep{Neal2012, campbell2021gradient} and higher-order approaches \citep{chen2015convergence, li2019stochastic} employ larger step sizes to improve stability. These approaches, however, still encounter difficulties in avoiding local traps, which is where advanced reMCMC methods help balance exploration and exploitation when navigating non-convex energy landscapes.

\section{DREXEL and DREAM with Binary Variables}
When the variable domain $\Theta$ is binary $\{0, 1\}^{\mathbf d}$, Algorithm \ref{alg:dream} can be further simplified. In this binary setting, the Hadamard product, denoted by $\odot$, simplifies several operations. This streamlined version demonstrates that both DREXEL and DREAM can be efficiently parallelized across CPUs and GPUs, which leads to reduced computational cost.

Binary variables are particularly advantageous in this context. The binary domain facilitates the use of efficient bitwise operations, which not only speed up the computation but also enable the algorithm to scale better in high-dimensional spaces. Moreover, the simplicity of the binary domain reduces algorithmic complexity, which can facilitate computational efficiency.

\section{DREXEL and DREAM with Categorical Variables}
We further examine how DREXEL and DREAM  can be formulated for categorical variables using one-hot vectors and ordinal integers. 

In \textbf{one-hot encoding}, each categorical variable $\btheta_i$ is represented as a vector in $\{0,1\}^N$ where exactly one element is 1, and the rest are 0. The update rule for one-hot encoded variables is given by:

$$\text{Categorical}\left[\text{Softmax}\left(\frac{1}{2\tau_k}\nabla U(\btheta_i^{(k)})_d\left(\theta_{i+1,d}^{(k)}-\theta_{i,d}^{(k)}\right) - \frac{\left\|\theta_{i+1,d}^{(k)}-\theta_{i, d}^{(k)}\right\|_2^2}{2\alpha_k}\right)\right], \quad k=1,2.$$

In this setting, the difference $\btheta_{i+1}^{(k)}-\btheta_{i}^{(k)}$ results in a vector with exactly two non-zero elements, which reflects a transition between categories.

For \textbf{ordinal variables}, where categories have a natural ordering, $\btheta_i$ can be represented as integers in $\{0, 1, \cdots, N-1\}$. The update rule becomes:

$$\text{Categorical}\left[\text{Softmax}\left(\frac{1}{2\tau_k}\nabla U(\btheta_i^{(k)})_d\left(\theta_{i+1,d}^{(k)}-\theta_{i,d}^{(k)}\right) - \frac{\left(\theta_{i+1,d}^{(k)}-\theta_{i, d}^{(k)}\right)^2}{2\alpha_k}\right)\right], \quad k=1,2.$$

Here, the scalar difference ($\theta_{i+1,d}^{(k)}-\theta_{i, d}^{(k)}$) captures the magnitude and direction of the transition between ordered categories. This representation leverages the ordering information to inform the proposal distribution more precisely.
\begin{algorithm}[!htbp]
  \caption{DREXEL or DREAM with Binary Variables.}\label{alg:binary}
{\textbf{Input} Step Sizes $\alpha_1$, $\alpha_2$, Temperatures $\tau_1$, $\tau_2$, and Swap Intensity $\rho > 0$.}\\
{\textbf{Input} Initial Samples $\boldsymbol \btheta_0^{(k)}\in \Theta$, $k=1, 2$.}
{\small
\begin{algorithmic}[1]
\State \textbf{For} $i=1,2,\cdots, I$ \textbf{do}
    \vspace{0.1em}
    \State \ \ \ \ \textcolor{darkred}{\textbf{Sampling Steps:}}
    \State \ \ \ \ \textbf{For} $k=0,1,2$ \textbf{do:}
    \State \ \ \ \ \ \ \ \ Compute $P_k(\btheta_i^{(k)}) = \displaystyle\frac{\exp\left(-\frac{1}{2\tau_k} \nabla U(\btheta_i^{(k)}) \odot (2\btheta_i^{(k)}-1) - \frac{1}{2\alpha_k \tau_k}\right)}{\exp\left(-\frac{1}{2\tau_k} \nabla U(\btheta_i^{(k)}) \odot (2\btheta_i^{(k)}-1) - \frac{1}{2\alpha_k \tau_k}\right) + 1}$
    \State \ \ \ \ \ \ \ \  Sample $\bm u \sim U([0,1]^{\mathbf d})$
    \State \ \ \ \ \ \ \ \  Set $\bm I_{k} \leftarrow \texttt{dim}(\bm u\le P_{k}(\btheta_i^{(k)}))$
    \State \ \ \ \ \ \ \ \  Set $\bomega^{(k)}\leftarrow \texttt{flipdim}(I_{k})$
    \State \ \ \ \ \textbf{End For} 
    \vspace{0.3em}
    \State \ \ \ \ \textcolor{darkgreen}{\textbf{MH Steps (for DREAM):}}
    \State \ \ \ \ \textbf{For} $k=1,2$ \textbf{do:}
    \State \ \ \ \ \ \ \ \ Compute $q_k(\bomega^{(k)}|\btheta_i^{(k)})=\Pi_{d\in \bm I_{k}}\mathbb P(\btheta_i^{(k)})_d\cdot \Pi_{d\notin \bm I_{k}}\left(1-\mathbb P(\btheta_i^{(k)})_d\right)$
    \State \ \ \ \ \ \ \ \ Compute $q_k(\btheta_i^{(k)}|\bomega^{(k)})=\Pi_{d\in \bm I_{k}}\mathbb P(\bomega^{(k)})_d\cdot \Pi_{d\notin \bm I_{k}}\left(1-\mathbb P(\bomega^{(k)})_d\right)$
    \State \ \ \ \ \ \ \ \ Compute $P_k(\bomega^{(k)}) = \displaystyle\frac{\exp\left(-\frac{1}{2\tau_k} \nabla U(\bomega^{(k)}) \odot (2\bomega^{(k)}-1) - \frac{1}{2\alpha_k \tau_k}\right)}{\exp\left(-\frac{1}{2\tau_k} \nabla U(\bomega^{(k)}) \odot (2\bomega^{(k)}-1) - \frac{1}{2\alpha_k \tau_k}\right) + 1}$
    \State \ \ \ \ \ \ \ \ Compute ${\mathcal{A}}(\bomega^{(k)}, \btheta_i^{(k)})$ follows \eqref{eq:mh_langevin}
    \State \ \ \ \ \ \ \ \ Generate a number $u\sim U[0,1]$
    \State \ \ \ \ \ \ \ \ Set $\boldsymbol{\btheta}^{(k)}_{i+1}\leftarrow \bomega^{(k)}$ if $u\leq {\mathcal{A}}$ else $\boldsymbol{\btheta}^{(k)}_{i+1}\leftarrow \boldsymbol \btheta^{(k)}_{i}$
    \State \ \ \ \ \textbf{End For} 
    \vspace{0.3em}
    \State\ \ \ \  \textcolor{darkblue}{\textbf{Swapping Steps:}}
    \State \ \ \ \ Generate a number $u\sim U[0,1]$.
    \State \ \ \ \ Compute $\tilde S(\boldsymbol \btheta^{(1)}_{i+1},\boldsymbol \btheta^{(2)}_{i+1})$ follows \eqref{eq:swap_history}
    \State \ \ \ \ Swap $\boldsymbol \btheta^{(1)}_{i+1}$ and $\boldsymbol \btheta^{(2)}_{i+1}$ if {$u\leq \rho\min\left\{1, \tilde S\right\}$}
    \vspace{0.1em}
\State \textbf{End For} \vspace{-0.2em}
\end{algorithmic}}
{\textbf{Output} Samples $\{\boldsymbol \btheta^{(1)}_i\}_{i=1}^{I}$.}\\ \vskip -0.10in
\end{algorithm}

\clearpage

\section{Theoretical Analysis}\label{appendix:sec:analysis}

To investigate the reversibility of the discrete replica exchange Langevin sampler, we build on the analysis of \citet{zhang2022langevin} and extend it to incorporate replica exchange. Specifically, we consider the transition kernels corresponding to the low- and high-temperature chains in the discrete setting:
\begin{equation*}
    \begin{split}
        \omega^{(1)}\sim q_1\left(\omega \mid \btheta^{(1)}\right) = \frac{\text{exp}\left[\frac{1}{2\tau_1}\nabla U(\btheta^{(1)})^\intercal\left(\omega-\btheta^{(1)}\right)-\frac{\left\|\omega-\btheta^{(1)}\right\|_2^2}{2\alpha_1}\right]}{Z_{\alpha_1}\left(\btheta^{(1)}\right)},\\
        % \propto  \text{Categorical}\left(\text{Softmax}\left(\frac{1}{\tau_1}\nabla U(\btheta^{(1)})_d(\theta_d'-\theta^{(1)}_d) - \frac{(\theta_d'-\theta^{(1)}_d)^2}{2\alpha_1}\right)\right)\\
        \omega^{(2)}\sim q_2\left(\omega \mid \btheta^{(2)}\right)= \frac{\text{exp}\left[\frac{1}{2\tau_2}\nabla U(\btheta^{(2)})^\intercal\left(\omega-\btheta^{(2)}\right)-\frac{\left\|\omega-\btheta^{(2)}\right\|_2^2}{2\alpha_2}\right]}{Z_{\alpha_2}\left(\btheta^{(2)}\right)},\\
        % \propto \text{Categorical}\left(\text{Softmax}\left(\frac{1}{\tau_2}\nabla U(\btheta^{(2)})_d(\theta_d'-\theta^{(2)}_d) - \frac{(\theta_d'-\theta^{(2)}_d)^2}{2\alpha_2}\right)\right),
    \end{split}
\end{equation*}
where each kernel $q_k$ uses a step size $\alpha_k$ and temperature $\tau_k$ for $k \in \{1,2\}$, and ${Z_{\alpha_k}\left(\btheta^{(k)}\right)}$ denotes the corresponding normalizing constant.

\subsection{Proof of Theorem \ref{theorem:weak_converge}}\label{appendix:subsec:balance}

\paragraph{Reversibility of the Discrete Replica Exchange Sampler}
\begin{proof}

We consider the joint transition kernel $q\left(\btheta_{i+1}^{(1)}, \btheta_{i+1}^{(2)} \mid \btheta_{i}^{(1)}, \btheta_{i}^{(2)}\right)$. Unlike the direct transition studied in \cite{zhang2022langevin}, we consider two possible transitions from $(\btheta_{i}^{(1)}, \btheta_{i}^{(2)})$ to $(\btheta_{i+1}^{(1)}, \btheta_{i+1}^{(2)})$. 
\begin{enumerate}
    \item With probability $1-S$, there is no chain swap, and the model parameter change from $(\btheta_{i}^{(1)},\btheta_{i}^{(2)})$ to $(\btheta_{i+1}^{(1)}, \btheta_{i+1}^{(2)})$ in the low-temperature sampler. Each replica evolves independently using its own Langevin-based transition kernel.
    \item With probability $S$, the two replicas exchange their roles and yield a cross-updated pair from $(\btheta_{i}^{(1)}, \btheta_{i}^{(2)})$ to $(\btheta_{i+1}^{(2)}, \btheta_{i+1}^{(1)})$.
\end{enumerate}
We recall the definition of the proposed swap function $S(\cdot,\cdot)$ in \eqref{eq:swap_history} as follows:

\begin{equation}\label{S_exact2}
% % \small
% \begin{split}
% S(\theta^{(1)},\theta^{(2)}):=e^{ \left(\frac{1}{\tau_2}-\frac{1}{\tau_1}\right)\left(U(\theta^{(1)})-U(\theta^{(2)})\right)}.
% \end{split}
S\left(\btheta_{i+1}^{(1)},\btheta_{i+1}^{(2)}\mid {\btheta_{i}^{(1)},\btheta_{i}^{(2)}}\right):=\min \left\{1, e^{ \left(\frac{1}{\tau_2}-\frac{1}{\tau_1}\right)\left[U\left(\btheta_{i+1}^{(1)}\right)+{U\left(\btheta_{i}^{(1)}\right)}-U\left(\btheta_{i+1}^{(2)}\right)-{U\left(\btheta_{i}^{(2)}\right)}\right]}\right\}.
\end{equation}

To simplify the notation, we denote 
\begin{equation}\label{eqn:swap-probability}
\begin{aligned}
\Delta U & := \left(\frac{1}{\tau_2}-\frac{1}{\tau_1}\right)\left[U\left(\btheta_{i+1}^{(1)}\right)+{U\left(\btheta_{i}^{(1)}\right)}-U\left(\btheta_{i+1}^{(2)}\right)-{U\left(\btheta_{i}^{(2)}\right)}\right], \\ 
S & := S\left(\btheta_{i+1}^{(1)},\btheta_{i+1}^{(2)}\mid {\btheta_{i}^{(1)},\btheta_{i}^{(2)}}\right) = \min \left\{1, e^{ \Delta U}\right\}
% & = \left(\frac{1}{\tau_2} - \frac{1}{\tau_1}\right) \Big[ \theta_{i+1}^{(1)\intercal} J \theta_{i+1}^{(1)} + \theta_i^{(1)\intercal} J \theta_i^{(1)} - \theta_{i+1}^{(2)\intercal} J \theta_{i+1}^{(2)} - \theta_i^{(2)\intercal} J \theta_i^{(2)} + \mathbf{b}^\intercal (\theta_{i+1}^{(1)} + \theta_i^{(1)} - \theta_{i+1}^{(2)} - \theta_i^{(2)}) \Big].
\end{aligned}
\end{equation}

The joint transition kernel from $(\btheta_{i}^{(1)}, \btheta_{i}^{(2)})$ to $(\btheta_{i+1}^{(1)}, \btheta_{i+1}^{(2)})$ can be written as:
\begin{equation}\label{eq:joint_transition}
    \begin{aligned}
& q\left(\btheta_{i+1}^{(1)},\btheta_{i+1}^{(2)}\big|\btheta_i^{(1)},\btheta_i^{(2)}\right) = \underbrace{\left(1 - S\right)
 q_1\left(\btheta_{i+1}^{(1)} \mid \btheta_i^{(1)}\right)
 q_2\left(\btheta_{i+1}^{(2)} \mid \btheta_i^{(2)}\right)}_{\text{no swap}} + \underbrace{S
 q_1\left(\btheta_{i+1}^{(2)} \mid \btheta_i^{(1)}\right)
 q_2\left(\btheta_{i+1}^{(1)} \mid \btheta_i^{(2)}\right)}_{\text{swap}}
\end{aligned}
\end{equation}
where the joint transition kernel in \eqref{eq:joint_transition} consists of two distinct cases.  The first term represents the case where no swap takes place between the low-temperature and high-temperature chains, while the second term quantifies the probability of a swap between the chains. Importantly, the marginal transition kernels $q_1$ (for the low-temperature chain) and $q_2$ (for the high-temperature chain) are independent, which directly leads to the formulation in \eqref{eq:joint_transition}.

We begin by recalling the discrete transition kernel used in our sampler. For a fixed configuration $\btheta_{i}$, the proposal distribution $q_{k}(\btheta_{i+1} \mid \btheta_{i})$ ($k=1,2$) at the $k$-th chain is defined via a first-order approximation of the target distribution, modulated by a quadratic regularizer scaled by $\alpha_k$. Specifically, we recall $q_{k}$ as:
\begin{equation*}
\begin{split}
q_{k}(\btheta_{i+1} \mid \btheta_{i}) &= \frac{
\exp\left( 
\frac{1}{2\tau_k} \nabla U(\btheta_{i})^\top (\btheta_{i+1} - \btheta_{i}) 
- \frac{1}{2\alpha_k} \|\btheta_{i+1} - \btheta_{i}\|^2 
\right)
}{
\sum_{x \in \Theta} 
\exp\left( 
\frac{1}{2\tau_k} \nabla U(\btheta_{i})^\top (x - \btheta_{i}) 
- \frac{1}{2\alpha_k} \|x - \btheta_{i}\|^2 
\right)
} \\ 
& = \frac{
\exp\left( 
\frac{1}{2\tau_k}(U(\btheta_{i+1}) - U(\btheta_{i})) 
- \frac{1}{2} (\btheta_{i+1} - \btheta_{i})^\top \left( \frac{1}{\alpha_k} I + \frac{1}{\tau_k} J \right) (\btheta_{i+1} - \btheta_{i})
\right)
}{
\sum_{x \in \Theta} 
\exp\left( 
\frac{1}{2\tau_k}(U(x) - U(\btheta_{i})) 
- \frac{1}{2} (x - \btheta_{i})^\top \left( \frac{1}{\alpha_k} I + \frac{1}{\tau_k} J \right) (x - \btheta_{i})
\right)
},
\end{split}
\end{equation*}
where the second equality holds by applying a first-order Taylor approximation of the energy function and regularizing it with a quadratic penalty. When the energy function is log-quadratic, i.e., 
$$
U(\btheta) = \btheta^\top J \btheta + \bm{b}^\top \btheta, 
$$
The gradient and energy difference admit closed-form expressions:
$$
\nabla U(\btheta) = 2J\btheta + \bm{b}, \quad 
U(\btheta_{i+1}) - U(\btheta_{i}) = \nabla U(\btheta_{i})^\top (\btheta_{i+1} - \btheta_{i}) + (\btheta_{i+1} - \btheta_{i})^\top J (\btheta_{i+1} - \btheta_{i}).
$$

We also recall that the marginal target distribution at the $k$-th replica is defined as 
$$
\pi_k(\btheta^{(k)}) \propto \exp\left[ \frac{1}{\tau_k} U(\btheta^{(k)}) \right],
$$
and the joint target distribution is 
$$
\pi(\btheta^{(1)}, \btheta^{(2)}) = \pi_1(\btheta^{(1)}) \pi_2(\btheta^{(2)}).
$$
To facilitate analysis, we define the auxiliary normalization function following \citet{zhang2022langevin}:
\begin{equation*}
Z_{\alpha_k}(\btheta^{(k)}) = \sum_{x \in \Theta} \exp\left[ \frac{1}{2} (U(x) - U(\btheta^{(k)})) - \frac{1}{2}(x - \btheta^{(k)})^\top \left( \frac{1}{\alpha_k} I + \frac{1}{\tau_k} J \right) (x - \btheta^{(k)}) \right], \quad k=1,2.
\end{equation*}

The intermediate target distribution is then defined as
$$
\tilde\pi(\btheta^{(1)}, \btheta^{(2)}) \propto Z_{\alpha_1}(\btheta^{(1)}) Z_{\alpha_2}(\btheta^{(2)}) \pi(\btheta^{(1)}, \btheta^{(2)}),
$$
which reduces to the original product-form target as $\alpha_k \to 0$. In the following, we show that the Markov transition kernel from $(\btheta_{i}^{(1)}, \btheta_{i}^{(2)})$ to $(\btheta_{i+1}^{(1)}, \btheta_{i+1}^{(2)})$ satisfies detailed balance with respect to $\tilde\pi$ under the log-quadratic structure of $U$.
% The intermediate distribution becomes $\tilde\pi\left(\btheta^{(1)}, \btheta^{(2)}\right) \propto Z_{\alpha_1}\left(\btheta^{(1)}\right) Z_{\alpha_2}\left(\btheta^{(2)}\right) \pi\left(\btheta^{(1)}, \btheta^{(2)}\right)$, which recovers the original target in the limit $\alpha_k \to 0$. We then proceed to demonstrate that the transition kernel between consecutive states from  $\left(\btheta_{i}^{(1)}, \btheta_{i}^{(2)}\right)$ to $\left(\btheta_{i+1}^{(1)}, \btheta_{i+1}^{(2)}\right)$ then preserves detailed balance with respect to $\tilde\pi\left(\btheta^{(1)}, \btheta^{(2)}\right)$ through the log-quadratic properties of the energy function $U(\btheta)$. 
We first decompose the product of $\tilde\pi\left(\btheta_i^{(1)}, \btheta_i^{(2)}\right)q\left(\btheta_{i+1}^{(1)}, \btheta_{i+1}^{(2)} \mid \btheta_i^{(1)}, \btheta_i^{(2)}\right)$ as follows:
\begin{equation}\label{eq:detailed-LHS}
    \begin{aligned}
        &\tilde\pi\left(\btheta_i^{(1)}, \btheta_i^{(2)}\right)q\left(\btheta_{i+1}^{(1)}, \btheta_{i+1}^{(2)} \mid \btheta_i^{(1)}, \btheta_i^{(2)}\right) \\
        = & \tilde\pi\left(\btheta_i^{(1)}, \btheta_i^{(2)}\right)\left[\left(1 - S\right)
 q_1\left(\btheta_{i+1}^{(1)} \mid \btheta_i^{(1)}\right)
 q_2\left(\btheta_{i+1}^{(2)} \mid \btheta_i^{(2)}\right) + S
 q_1\left(\btheta_{i+1}^{(2)} \mid \btheta_i^{(1)}\right)
 q_2\left(\btheta_{i+1}^{(1)} \mid \btheta_i^{(2)}\right) \right] \\
        = & \frac{Z_{\alpha_1}\left(\btheta_{i}^{(1)}\right) Z_{\alpha_2}\left(\btheta_{i}^{(2)}\right)\pi\left(\btheta_{i}^{(1)}, \btheta_{i}^{(2)}\right)}{D}\Big[\left(1 - S\right)\cdot T_1 + S\cdot T_2 \Big], \\
        % = & \frac{\exp \left[\frac{1}{2}\left(U\left(\btheta_{i+1}^{(2)}\right)+U(\btheta_{i}^{(2)})\right)-\left(\btheta_{i+1}^{(2)}-\btheta_{i}^{(2)}\right)^{\intercal}\left(\frac{1}{2 \alpha} \mathbf I +\frac{1}{2} \bJ\right)\left(\btheta_{i+1}^{(2)}-\btheta_{i}^{(2)}\right)\right]}{Z \cdot \sum_{\bm x} Z({\bm x}) \pi_\alpha({\bm x})},
    \end{aligned}
\end{equation}
where $D:= \sum_{x\in\Theta}\sum_{y\in\Theta}Z_{\alpha_1}(x)Z_{\alpha_2}(y)\pi(x,y)$. We continue to analyze $T_1$ and $T_2$. For $T_1$ we have
\begin{equation}\label{eq:T1}
    \begin{aligned}
        T_1 & = q_1\left(\btheta_{i+1}^{(1)} \mid \btheta_i^{(1)}\right)
 q_2\left(\btheta_{i+1}^{(2)} \mid \btheta_i^{(2)}\right) \\ 
 & = \frac{\exp \left[\frac{1}{2\tau_1}\left(U\left(\btheta_{i+1}^{(1)}\right)-U\left(\btheta_{i}^{(1)}\right)\right)-\left(\btheta_{i+1}^{(1)}-\btheta_{i}^{(1)}\right)^{\intercal}\left(\frac{1}{2 \alpha_1} \mathbf I +\frac{1}{2\tau_1} \bJ\right)\left(\btheta_{i+1}^{(1)}-\btheta_{i}^{(1)}\right)\right]}{Z_{\alpha_1}  \left(\btheta^{(1)} \right)} \\ 
 & \qquad \cdot\frac{\exp \left[\frac{1}{2\tau_2}\left(U\left(\btheta_{i+1}^{(2)}\right)-U\left(\btheta_{i}^{(2)}\right)\right)-\left(\btheta_{i+1}^{(2)}-\btheta_{i}^{(2)}\right)^{\intercal}\left(\frac{1}{2 \alpha_2} \mathbf I +\frac{1}{2\tau_2} \bJ\right)\left(\btheta_{i+1}^{(2)}-\btheta_{i}^{(2)}\right)\right]}{Z_{\alpha_2}  \left(\btheta^{(2)} \right)}. \\
 % & = \frac{\exp \left[\frac{1}{2}\left(U\left(\btheta_{i+1}^{(1)}\right)+U(\btheta_{i}^{(1)})+U\left(\btheta_{i+1}^{(2)}\right)+U(\btheta_{i}^{(2)})\right)\right]}{Z \sum_{\btheta^\prime}\sum_{\btheta} Z({\btheta})Z({\btheta^\prime}) \pi_\alpha({\btheta}, \btheta^\prime)} \\ 
 % & \qquad \cdot \frac{\exp \left[-\left(\btheta_{i+1}^{(1)}-\btheta_{i}^{(1)}+\btheta_{i+1}^{(2)}-\btheta_{i}^{(2)}\right)^{\intercal}\left(\frac{1}{2 \alpha} \mathbf I +\frac{1}{2} \bJ\right)\left(\btheta_{i+1}^{(1)}-\btheta_{i}^{(1)}+\btheta_{i+1}^{(2)}-\btheta_{i}^{(2)}\right)\right]}{Z \sum_{\btheta^\prime}\sum_{\btheta} Z({\btheta})Z({\btheta^\prime}) \pi_\alpha({\btheta}, \btheta^\prime)}.
    \end{aligned}
\end{equation}
Similarly, we can decompose $T_2$ as follows:
\begin{equation}\label{eq:T2}
    \begin{aligned}
        T_2 & = q_1\left(\btheta_{i+1}^{(2)} \mid \btheta_i^{(1)}\right)
 q_2\left(\btheta_{i+1}^{(1)} \mid \btheta_i^{(2)}\right)\\
 & = \frac{\exp \left[\frac{1}{2\tau_1}\left(U\left(\btheta_{i+1}^{(2)}\right)-U\left(\btheta_{i}^{(1)}\right)\right)-\left(\btheta_{i+1}^{(2)}-\btheta_{i}^{(1)}\right)^{\intercal}\left(\frac{1}{2 \alpha_1} \mathbf I +\frac{1}{2\tau_1} \bJ\right)\left(\btheta_{i+1}^{(2)}-\btheta_{i}^{(1)}\right)\right]}{Z_{\alpha_1}  \left(\btheta^{(1)} \right)} \\ 
 & \qquad \cdot\frac{\exp \left[\frac{1}{2\tau_2}\left(U\left(\btheta_{i+1}^{(1)}\right)-U\left(\btheta_{i}^{(2)}\right)\right)-\left(\btheta_{i+1}^{(1)}-\btheta_{i}^{(2)}\right)^{\intercal}\left(\frac{1}{2 \alpha_2} \mathbf I +\frac{1}{2\tau_2} \bJ\right)\left(\btheta_{i+1}^{(1)}-\btheta_{i}^{(2)}\right)\right]}{Z_{\alpha_2}  \left(\btheta^{(2)} \right)}. \\
    \end{aligned}
\end{equation}

Incorporating \eqref{eq:T1}-\eqref{eq:T2} and the definition of $\pi(\btheta^{(1)}, \btheta^{(2)})$, we reform \eqref{eq:detailed-LHS} as:
\begin{equation}\label{eq:detailed-LHS2}
    \begin{split}
        \tilde\pi\left(\btheta_i^{(1)}, \btheta_i^{(2)}\right)q\left(\btheta_{i+1}^{(1)}, \btheta_{i+1}^{(2)} \mid \btheta_i^{(1)}, \btheta_i^{(2)}\right) & = \frac{Z_{\alpha_1}\left(\btheta_{i}^{(1)}\right) Z_{\alpha_2}\left(\btheta_{i}^{(2)}\right)\pi\left(\btheta_{i}^{(1)}, \btheta_{i}^{(2)}\right)}{D}\Big[\left(1 - S\right)\cdot T_1 + S\cdot T_2 \Big] \\ 
        & = \frac{1}{D}\left[\left(1 - S\right)\cdot T_1^\prime + S\cdot T_2^\prime \right], \\ 
    \end{split}
\end{equation}
where we denote here $T_1^\prime$ and $T_1^\prime$ can be written as:
\begin{equation*}
    \begin{split}
        T_1^\prime & = \exp \left[\frac{1}{2\tau_1}\left(U\left(\btheta_{i+1}^{(1)}\right)+U\left(\btheta_{i}^{(1)}\right)\right)-\left(\btheta_{i+1}^{(1)}-\btheta_{i}^{(1)}\right)^{\intercal}\left(\frac{1}{2 \alpha_1} \mathbf I +\frac{1}{2\tau_1} \bJ\right)\left(\btheta_{i+1}^{(1)}-\btheta_{i}^{(1)}\right)\right] \\ 
        & \qquad \cdot\exp \left[\frac{1}{2\tau_2}\left(U\left(\btheta_{i+1}^{(2)}\right)+U\left(\btheta_{i}^{(2)}\right)\right)-\left(\btheta_{i+1}^{(2)}-\btheta_{i}^{(2)}\right)^{\intercal}\left(\frac{1}{2 \alpha_2} \mathbf I +\frac{1}{2\tau_2} \bJ\right)\left(\btheta_{i+1}^{(2)}-\btheta_{i}^{(2)}\right)\right]. \\
        T_2^\prime & = \exp \left[\frac{1}{2\tau_1}\left(U\left(\btheta_{i+1}^{(2)}\right)+U\left(\btheta_{i}^{(1)}\right)\right)-\left(\btheta_{i+1}^{(2)}-\btheta_{i}^{(1)}\right)^{\intercal}\left(\frac{1}{2 \alpha_1} \mathbf I +\frac{1}{2\tau_1} \bJ\right)\left(\btheta_{i+1}^{(2)}-\btheta_{i}^{(1)}\right)\right] \\ 
        & \qquad \cdot\exp \left[\frac{1}{2\tau_2}\left(U\left(\btheta_{i+1}^{(1)}\right)+U\left(\btheta_{i}^{(2)}\right)\right)-\left(\btheta_{i+1}^{(1)}-\btheta_{i}^{(2)}\right)^{\intercal}\left(\frac{1}{2 \alpha_2} \mathbf I +\frac{1}{2\tau_2} \bJ\right)\left(\btheta_{i+1}^{(1)}-\btheta_{i}^{(2)}\right)\right].\\
    \end{split}
\end{equation*}
%-------------------------------------------------------------
% Detailed Balance Derivation for Replica Exchange
%-------------------------------------------------------------

% Assume the following packages are loaded in the preamble:
% \usepackage{amsmath, amssymb}

To verify that the coupled Markov chain satisfies detailed balance with respect to the intermediate distribution $\tilde\pi$, we aim to show that $\forall \btheta_{i}^{(1)}, \btheta_{i}^{(2)}, \btheta_{i+1}^{(1)}, \btheta_{i+1}^{(2)}\in \Theta$ and sampled from Algorithm \ref{alg:dream}, the following equality holds:
\begin{equation}
\tilde\pi \left(\btheta_{i}^{(1)}, \btheta_{i}^{(2)} \right)
q \left(\btheta_{i+1}^{(1)}, \btheta_{i+1}^{(2)} \mid \btheta_{i}^{(1)}, \btheta_{i}^{(2)} \right)
=
\tilde\pi \left(\btheta_{i+1}^{(1)}, \btheta_{i+1}^{(2)} \right)
q \left(\btheta_{i}^{(1)}, \btheta_{i}^{(2)} \mid \btheta_{i+1}^{(1)}, \btheta_{i+1}^{(2)} \right).
\label{eq:detailed-balance-full}
\end{equation}

The reverse transition on the right-hand side of \eqref{eq:detailed-balance-full} can be decomposed as:
\begin{equation}\label{eq:detailed-RHS}
    \begin{aligned}
    &\textcolor{white}{=}\tilde\pi \left(\btheta_{i+1}^{(1)}, \btheta_{i+1}^{(2)} \right)
    \cdot q \left(\btheta_{i}^{(1)}, \btheta_{i}^{(2)} \mid \btheta_{i+1}^{(1)}, \btheta_{i+1}^{(2)} \right) \\
    &= \frac{Z_{\alpha_1}\left(\btheta_{i+1}^{(1)}\right) Z_{\alpha_2}\left(\btheta_{i+1}^{(2)}\right)
    \pi\left(\btheta_{i+1}^{(1)}, \btheta_{i+1}^{(2)}\right)}{D}\\ 
    & \qquad \cdot\left[ (1 - S') q_1 \left(\btheta_{i}^{(1)} \mid \btheta_{i+1}^{(1)} \right) q_2 \left(\btheta_{i}^{(2)} \mid \btheta_{i+1}^{(2)} \right) + S' q_1 \left(\btheta_{i}^{(2)} \mid \btheta_{i+1}^{(1)} \right) q_2 \left(\btheta_{i}^{(1)} \mid \btheta_{i+1}^{(2)} \right) \right],\\
    & = \frac{Z_{\alpha_1}\left(\btheta_{i+1}^{(1)}\right) Z_{\alpha_2}\left(\btheta_{i+1}^{(2)}\right)
    \pi\left(\btheta_{i+1}^{(1)}, \btheta_{i+1}^{(2)}\right)}{D}\Big[\left(1 - S'\right)\cdot T_3 + S'\cdot T_4 \Big] \\   
    & = \frac{1}{D}\Big[\left(1 - S'\right)\cdot T_3^\prime + S'\cdot T_4^\prime \Big]
    \end{aligned}
\end{equation}
where $S'$ is the reverse swap probability, which satisfies $S' = S$ due to symmetry of the energy differences in \eqref{S_exact2}. Subsequently, we express $T_3$, $T_4$, $T_3^\prime$ and $T_4^\prime$ as follows:
\begin{equation}\label{eq:T3T4}
    \begin{aligned}
        T_3 & = \frac{\exp \left[\frac{1}{2\tau_1}\left(U\left(\btheta_{i}^{(1)}\right)-U\left(\btheta_{i+1}^{(1)}\right)\right)-\left(\btheta_{i}^{(1)}-\btheta_{i+1}^{(1)}\right)^{\intercal}\left(\frac{1}{2 \alpha_1} \mathbf I +\frac{1}{2\tau_1} \bJ\right)\left(\btheta_{i}^{(1)}-\btheta_{i+1}^{(1)}\right)\right]}{Z_{\alpha_1}  \left(\btheta^{(1)} \right)} \\ 
 & \qquad \cdot\frac{\exp \left[\frac{1}{2\tau_2}\left(U\left(\btheta_{i}^{(2)}\right)-U\left(\btheta_{i+1}^{(2)}\right)\right)-\left(\btheta_{i}^{(2)}-\btheta_{i+1}^{(2)}\right)^{\intercal}\left(\frac{1}{2 \alpha_2} \mathbf I +\frac{1}{2\tau_2} \bJ\right)\left(\btheta_{i}^{(2)}-\btheta_{i+1}^{(2)}\right)\right]}{Z_{\alpha_2}  \left(\btheta^{(2)} \right)}, \\
        T_4 & = \frac{\exp \left[\frac{1}{2\tau_1}\left(U\left(\btheta_{i}^{(2)}\right)-U\left(\btheta_{i+1}^{(1)}\right)\right)-\left(\btheta_{i}^{(2)}-\btheta_{i+1}^{(1)}\right)^{\intercal}\left(\frac{1}{2 \alpha_1} \mathbf I +\frac{1}{2\tau_1} \bJ\right)\left(\btheta_{i}^{(2)}-\btheta_{i+1}^{(1)}\right)\right]}{Z_{\alpha_1}  \left(\btheta^{(1)} \right)} \\ 
 & \qquad \cdot\frac{\exp \left[\frac{1}{2\tau_2}\left(U\left(\btheta_{i}^{(1)}\right)-U\left(\btheta_{i+1}^{(2)}\right)\right)-\left(\btheta_{i}^{(1)}-\btheta_{i+1}^{(2)}\right)^{\intercal}\left(\frac{1}{2 \alpha_2} \mathbf I +\frac{1}{2\tau_2} \bJ\right)\left(\btheta_{i}^{(1)}-\btheta_{i+1}^{(2)}\right)\right]}{Z_{\alpha_2}  \left(\btheta^{(2)} \right)}, \\
         T_3^\prime & = \exp \left[\frac{1}{2\tau_1}\left(U\left(\btheta_{i}^{(1)}\right)\right)+U\left(\btheta_{i+1}^{(1)}\right)-\left(\btheta_{i}^{(1)}-\btheta_{i+1}^{(1)}\right)^{\intercal}\left(\frac{1}{2 \alpha_1} \mathbf I +\frac{1}{2\tau_1} \bJ\right)\left(\btheta_{i}^{(1)}-\btheta_{i+1}^{(1)}\right)\right] \\ 
        & \qquad \cdot\exp \left[\frac{1}{2\tau_2}\left(U\left(\btheta_{i}^{(2)}\right)\right)+U\left(\btheta_{i+1}^{(2)}\right)-\left(\btheta_{i}^{(2)}-\btheta_{i+1}^{(2)}\right)^{\intercal}\left(\frac{1}{2 \alpha_2} \mathbf I +\frac{1}{2\tau_2} \bJ\right)\left(\btheta_{i}^{(2)}-\btheta_{i+1}^{(2)}\right)\right]. \\
        T_4^\prime & = \exp \left[\frac{1}{2\tau_1}\left(U\left(\btheta_{i}^{(1)}\right)\right)+U\left(\btheta_{i+1}^{(2)}\right)-\left(\btheta_{i}^{(1)}-\btheta_{i+1}^{(2)}\right)^{\intercal}\left(\frac{1}{2 \alpha_1} \mathbf I +\frac{1}{2\tau_1} \bJ\right)\left(\btheta_{i}^{(1)}-\btheta_{i+1}^{(2)}\right)\right] \\ 
        & \qquad \cdot\exp \left[\frac{1}{2\tau_2}\left(U\left(\btheta_{i}^{(2)}\right)\right)+U\left(\btheta_{i+1}^{(1)}\right)-\left(\btheta_{i}^{(2)}-\btheta_{i+1}^{(1)}\right)^{\intercal}\left(\frac{1}{2 \alpha_2} \mathbf I +\frac{1}{2\tau_2} \bJ\right)\left(\btheta_{i}^{(2)}-\btheta_{i+1}^{(1)}\right)\right].\\
    \end{aligned}
\end{equation}
which we observe that $T_3^\prime=T_1^\prime$ and $T_4^\prime=T_2^\prime$ satisfy due to the symmetry. Since the forward and reverse terms in \eqref{eq:detailed-LHS2} and \eqref{eq:detailed-RHS} are symmetric under
$(\btheta_{i}^{(1)}, \btheta_{i}^{(2)}) \leftrightarrow (\btheta_{i+1}^{(1)}, \btheta_{i+1}^{(2)})$,
We conclude that detailed balance holds:
$$
\tilde\pi \left(\btheta_{i}^{(1)}, \btheta_{i}^{(2)} \right)
q \left(\btheta_{i+1}^{(1)}, \btheta_{i+1}^{(2)} \mid \btheta_{i}^{(1)}, \btheta_{i}^{(2)} \right)
=
\tilde\pi \left(\btheta_{i+1}^{(1)}, \btheta_{i+1}^{(2)} \right)
q \left(\btheta_{i}^{(1)}, \btheta_{i}^{(2)} \mid \btheta_{i+1}^{(1)}, \btheta_{i+1}^{(2)} \right).
$$
Therefore, the Markov chain is reversible with respect to the intermediate distribution $\tilde\pi$.
\end{proof}

\paragraph{Weak Convergence to the Target Distribution}
\begin{proof}
We continue to show that the intermediate distribution $\tilde\pi(\btheta^{(1)}, \btheta^{(2)})$ converges weakly to the target distribution $\pi(\btheta^{(1)}, \btheta^{(2)})$ as the step sizes $\alpha_1, \alpha_2 \to 0$. Recall the form of the correction term for the chains $k = 1, 2$:
$$
Z_{\alpha_k}(\btheta) =
\sum_{x \in \Theta}
\exp\left[
\frac{1}{2} \left(U(x) - U(\btheta)\right)
-
\frac{1}{2}(x - \btheta)^\top
\left( \frac{1}{\alpha_k} I + \frac{1}{\tau_k} J \right)
(x - \btheta)
\right].
$$

As $\alpha_k \to 0$, the quadratic form is dominated by the term $\frac{1}{\alpha_k} I$. For any $x \neq \btheta$, we have
$$
(x - \btheta)^\top\left( \frac{1}{\alpha_k} I + \frac{1}{\tau_k} J \right)(x - \btheta)
\geq \frac{1}{\alpha_k} \|x - \btheta\|_2^2 \to \infty.
$$
Hence, $\exp\left[-\frac{1}{2}(x - \btheta)^\top(\cdot)(x - \btheta)\right] \to 0$ for $x \neq \btheta$. For the case $x = \btheta$, the exponent simplifies to $0$ and thus contributes exactly $1$. Therefore, we can conclude $
\lim_{\alpha_k \to 0} Z_{\alpha_k}(\btheta) = 1.$. Subsequently, we further recall that the full joint normalization constant is
$$
D = \sum_{\boldsymbol x \in \Theta} \sum_{\boldsymbol y \in \Theta}
Z_{\alpha_1}(\boldsymbol x)\, Z_{\alpha_2}(\boldsymbol y)\, \pi(\boldsymbol x, \boldsymbol y).
$$
Since $Z_{\alpha_k}(\cdot) \to 1$ point-wise, we have
$
\lim_{\alpha_1, \alpha_2 \to 0} D = \sum_{\boldsymbol x, \boldsymbol y \in \Theta} \pi(\boldsymbol x, \boldsymbol y) = 1.$

The intermediate distribution is defined as
$$
\tilde\pi(\btheta^{(1)}, \btheta^{(2)}) =
\frac{Z_{\alpha_1}(\btheta^{(1)}) Z_{\alpha_2}(\btheta^{(2)}) \pi(\btheta^{(1)}, \btheta^{(2)})}{D}.
$$
Taking the limit on both sides, we can further obtain:
$$
\lim_{\alpha_1, \alpha_2 \to 0} \tilde\pi(\btheta^{(1)}, \btheta^{(2)})
= \frac{\pi(\btheta^{(1)}, \btheta^{(2)})}{1} = \pi(\btheta^{(1)}, \btheta^{(2)}),
$$
which implies that the stationary distribution $\tilde\pi$ converges point-wisely to the target $\pi$ on the discrete space $\Theta \times \Theta$.

Since both $\tilde\pi$ and $\pi$ are discrete probability mass functions over a finite state space, point-wise convergence implies convergence in total variation:
$$
\lim_{\alpha_1, \alpha_2 \to 0}
\sum_{(\btheta^{(1)}, \btheta^{(2)}) \in \Theta^2}
\left| \tilde\pi(\btheta^{(1)}, \btheta^{(2)}) - \pi(\btheta^{(1)}, \btheta^{(2)}) \right| = 0.
$$
By Scheff\'e's Lemma in \citet{billingsley2017probability}, this implies $\tilde\pi$ converges weakly to the target $ \pi$.

Finally, by the Dominated Convergence Theorem in \citet{folland1999real}, for any bounded function \(f: \Theta \times \Theta \to \mathbb{R}\), we have
$$
\lim_{\alpha_1, \alpha_2 \to 0}
\sum_{(\btheta^{(1)}, \btheta^{(2)}) \in \Theta^2}
f(\btheta^{(1)}, \btheta^{(2)})\, \tilde\pi(\btheta^{(1)}, \btheta^{(2)})
=
\sum_{(\btheta^{(1)}, \btheta^{(2)}) \in \Theta^2}
f(\btheta^{(1)}, \btheta^{(2)})\, \pi(\btheta^{(1)}, \btheta^{(2)}).
$$
This completes the proof of weak convergence.
\end{proof}

\subsection{Proof of Theorem \ref{thm:convergence2}}

We follow an argument similar to that in \citet[Proposition 3]{diaconis1991geometric}. It is straightforward to verify that $q(\btheta\mid\btheta') > 0$ for all $\btheta, \btheta' \in \Theta \times \Theta$ since the density functions are exponential. Therefore, the chain is irreducible. It has also been established that $q$ is reversible with respect to $\tilde\pi$, which implies
$$
\tilde\pi(\btheta) q^n(\btheta'\mid\btheta) = \tilde\pi(\btheta') q^n(\btheta\mid\btheta').
$$
This leads to the identity
$$
\frac{q^n(\btheta'\mid \btheta)}{\tilde\pi(\btheta')} = \frac{q^n(\btheta\mid\btheta')}{\tilde\pi(\btheta)}.
$$
Multiplying both sides by $q^n(\btheta'\mid\btheta)$ and summing over $\btheta'$, we obtain
\begin{equation}\label{eqn:q2nidentity}
\sum_{\btheta'} \frac{(q^n(\btheta'\mid\btheta))^2}{\tilde\pi(\btheta')} = \sum_{\btheta'} \frac{q^n(\btheta\mid\btheta') q^n(\btheta'\mid\btheta)}{\tilde\pi(\btheta)} = \frac{q^{2n}(\btheta\mid\btheta)}{\tilde\pi(\btheta)}.
\end{equation}

Using the fact that total variation distance is half the $L^1$ distance, we estimate:
\begin{equation}\label{eqn:qnestimate}
\begin{aligned}
\|q^n(\cdot\mid\btheta) - \tilde\pi(\cdot)\|_{\mathrm{TV}}^2 
&= \left( \frac{1}{2} \sum_{\btheta'} \left| q^n(\btheta'\mid\btheta) - \tilde\pi(\btheta') \right| \right)^2 \\
&= \left( \frac{1}{2} \sum_{\btheta'} \sqrt{\tilde\pi(\btheta')} 
\frac{\left| q^n(\btheta'\mid\btheta) - \tilde\pi(\btheta') \right|}{\sqrt{\tilde\pi(\btheta')}} \right)^2 \\
&\le \frac{1}{4} \left( \sum_{\btheta'} \tilde\pi(\btheta') \right) 
\left( \sum_{\btheta'} \frac{\left| q^n(\btheta'\mid\btheta) - \tilde\pi(\btheta') \right|^2}{\tilde\pi(\btheta')} \right) \\
&= \frac{1}{4} \sum_{\btheta'} \frac{(q^n(\btheta'\mid\btheta))^2 - 2q^n(\btheta'\mid\btheta)\tilde\pi(\btheta') + (\tilde\pi(\btheta'))^2}{\tilde\pi(\btheta')} \\
&= \frac{1}{4} \sum_{\btheta'} \left( \frac{(q^n(\btheta'\mid\btheta))^2}{\tilde\pi(\btheta')} - 2q^n(\btheta'\mid\btheta) + \tilde\pi(\btheta') \right) \\
&= \frac{1}{4} \left( \sum_{\btheta'} \frac{(q^n(\btheta'\mid\btheta))^2}{\tilde\pi(\btheta')} - 1 \right) 
= \frac{1}{4} \left( \frac{q^{2n}(\btheta\mid\btheta)}{\tilde\pi(\btheta)} - 1 \right),
\end{aligned}
\end{equation}
where we used the Cauchy--Schwarz inequality and identity \eqref{eqn:q2nidentity} in the final step.

Now, let $D$ be the diagonal matrix with entries $\sqrt{\tilde\pi(\cdot)}$. Then the matrix $DqD^{-1}$ has entries
$$
(DqD^{-1})_{\btheta \btheta'} = \sqrt{\frac{\tilde\pi(\btheta)}{\tilde\pi(\btheta')}} q(\btheta'\mid\btheta),
$$
which implies that $DqD^{-1}$ is symmetric. Hence, it admits an orthogonal diagonalization:
$$
DqD^{-1} = Q \tilde{q} Q^\top,
$$
for some orthogonal matrix $Q$ and diagonal matrix $\tilde{q}$ whose entries are the eigenvalues of $q$, denoted as $1 = \lambda_0 > \lambda_1 \ge \cdots \ge \lambda_{N^{2\mathbf{d}} - 1} \ge -1$.

Consequently, we have
$$
q^{2n} = D^{-1} Q \tilde{q}^{2n} Q^\top D,
$$
and
$$
q^{2n}(\btheta\mid\btheta) = \tilde\pi(\btheta) + \sum_{\btheta' \neq 0} \lambda_{\btheta'}^{2n} Q_{\btheta \btheta'}^2.
$$

Substituting this into the earlier bound \eqref{eqn:qnestimate}, we obtain
\begin{equation*}
\begin{aligned}
\|q^n(\cdot\mid\btheta) - \tilde\pi(\cdot)\|_{\mathrm{TV}}^2 
&\le \frac{1}{4} \left( \frac{\tilde\pi(\btheta) + \sum_{\btheta'\neq 0} \lambda_{\btheta'}^{2n} Q_{\btheta \btheta'}^2}{\tilde\pi(\btheta)} - 1 \right) \\
&= \frac{1}{4\tilde\pi(\btheta)} \sum_{\btheta'\neq 0} \lambda_{\btheta'}^{2n} Q_{\btheta \btheta'}^2 \\
&\le \frac{1}{4\tilde\pi(\btheta)} \lambda_*^{2n}.
\end{aligned}
\end{equation*}

\section{Additional Experimental Results}\label{appendix:sec:exp}
% \haoyang{Please refer to \cite{dai2020learning} for experimental setup}

The experiments were run on a server featuring an Intel(R) Core(TM) i9-14900K processor, RTX 4090 GPUs, and 128 GB DDR4 memory.

\subsection{Sampling from 2D Synthetic Energies}\label{appendix:subsec_synthetic}

To evaluate the efficiency of MCMC algorithms in exploring non-convex discrete energy landscapes, we consider a set of energy functions that present varying degrees of complexity and multimodality. These functions are designed to test the algorithms' ability to navigate challenging landscapes characterized by multiple local minima, sharp ridges, and disconnected modes.

\textbf{Wave} energy function is a periodic sinusoidal surface with alternating peaks and valleys:
\begin{equation*}\label{eq:energy_wave}
    U(x, y) = \sin(3 x) \sin(3 y).
\end{equation*}
Its highly rugged landscape arises from repeated oscillations, which makes it a suitable test for an algorithm's ability to navigate sharp and oscillating energy contours without getting trapped in local minima.

\textbf{Eight Gaussian} energy function \citep{dai2020learning} consists of eight equally spaced Gaussian components arranged in a circular pattern:

\begin{equation*}\label{eq:energy_8gauss}
    U(x, y) = \sum_{i=1}^8 \frac{-\left[(x-x_i)^2 + (y-y_i)^2\right]}{2\sigma^2},
\end{equation*}
where the centers $(x_i, y_i)$ are located at $(\pm1, 0)$, $(0, \pm1)$, and $\left(\pm\frac{\sqrt{2}}{2}, \pm\frac{\sqrt{2}}{2}\right)$, with $\sigma = 1.0$ for all modes. This function presents multiple well-separated basins of attraction, testing the algorithm's ability to explore disconnected modes effectively.

\textbf{Moon} energy function creates a complex, asymmetric landscape resembling a crescent shape:
\begin{equation*}\label{eq:energy_moon}
    U(x, y) = -\frac{1}{10}y^4 - \frac{1}{2}\left(4x - y^2 + \frac{24}{5}\right)^2.
\end{equation*}
With deep valley and steep ridge features, this non-convex structure evaluates the algorithm's capacity to explore non-uniform, curved regions and traverse narrow channels between high-energy barriers.

\textbf{Two Moons} energy function describes a landscape with two prominent crescent-shaped modes separated by a low-energy region:
\begin{equation*}\label{eq:energy_twomoons}
    U(x, y) = -\frac{2}{25}\left({x^2 + y^2 - 2}\right)^2 + \log\left[e^{-\frac{1}{2}\left(\frac{5x - 4}{4}\right)^2} + e^{-\frac{1}{2}\left(\frac{5x + 4}{4}\right)^2}\right]
\end{equation*}
This function challenges MCMC methods to jump between distinct modes, and tests their efficiency in exploring multi-modal distributions where modes are not directly connected.

\textbf{Twist} energy function represents a twisted sinusoidal landscape where energy levels change smoothly along a sinusoidal curve:
\begin{equation*}\label{eq:energy_twist}
    U(x, y) = -\frac{1}{2}\left[y-\sin\left(\frac{\pi x}{2}\right)\right]^2.
\end{equation*}
The narrow, twisted valleys require careful gradient following, which tests the algorithm's ability to sample from highly structured, nonlinear regions of the energy landscape.

\textbf{Flower} energy function combines radial symmetry with angular variations, forming a complex landscape with a petal-like structure:
\begin{equation*}\label{eq:energy_flower}
    U(x, y) = \sin{\left(\sqrt{ x^2 + y^2}\right)}+ \cos\left(5\tan \left(\frac{y}{x}\right)\right).
\end{equation*}
With multiple local minima and ridges radiating from the center, this intricate landscape tests the algorithm's capability to explore multi-modal, rotationally symmetric energy landscapes with sharp transitions between regions.

A detailed hyperparameter setting of DREAM for each task is shown in Table \ref{tab:synthetic_parameters}

\begin{table}[!htbp]
\centering\caption{Hyperparameters used in 2d synthetic simulations. For each dataset, the low-temperature chain step size ($\alpha_1$), high-temperature chain step size ($\alpha_2$), and high temperatures ($\tau_2$) are reported.}\label{tab:synthetic_parameters}
\begin{tabular}{ccccccccc}
\toprule
      & wave  & 8gaussians  & 16gaussians   & moon  & 2moons & twist & flower \\ \midrule
$\alpha_1$ & 0.025 & 0.025    & 0.025       & 0.025 & 0.025  & 0.025 & 0.025  \\
$\alpha_2$ & 0.055 & 0.055    & 0.052       & 0.065 & 0.065  & 0.065 & 0.065  \\
$\tau_1$   & 1.0   & 1.0      & 1.0         & 1.0   & 1.0    & 1.0   & 1.0  \\
$\tau_2$   & 5.0   & 5.0      & 2.0         & 5.0   & 5.0    & 5.0   & 5.0       \\ \bottomrule
\end{tabular}
\end{table}

\begin{figure*}[t]

\includegraphics[width=1.0\textwidth]{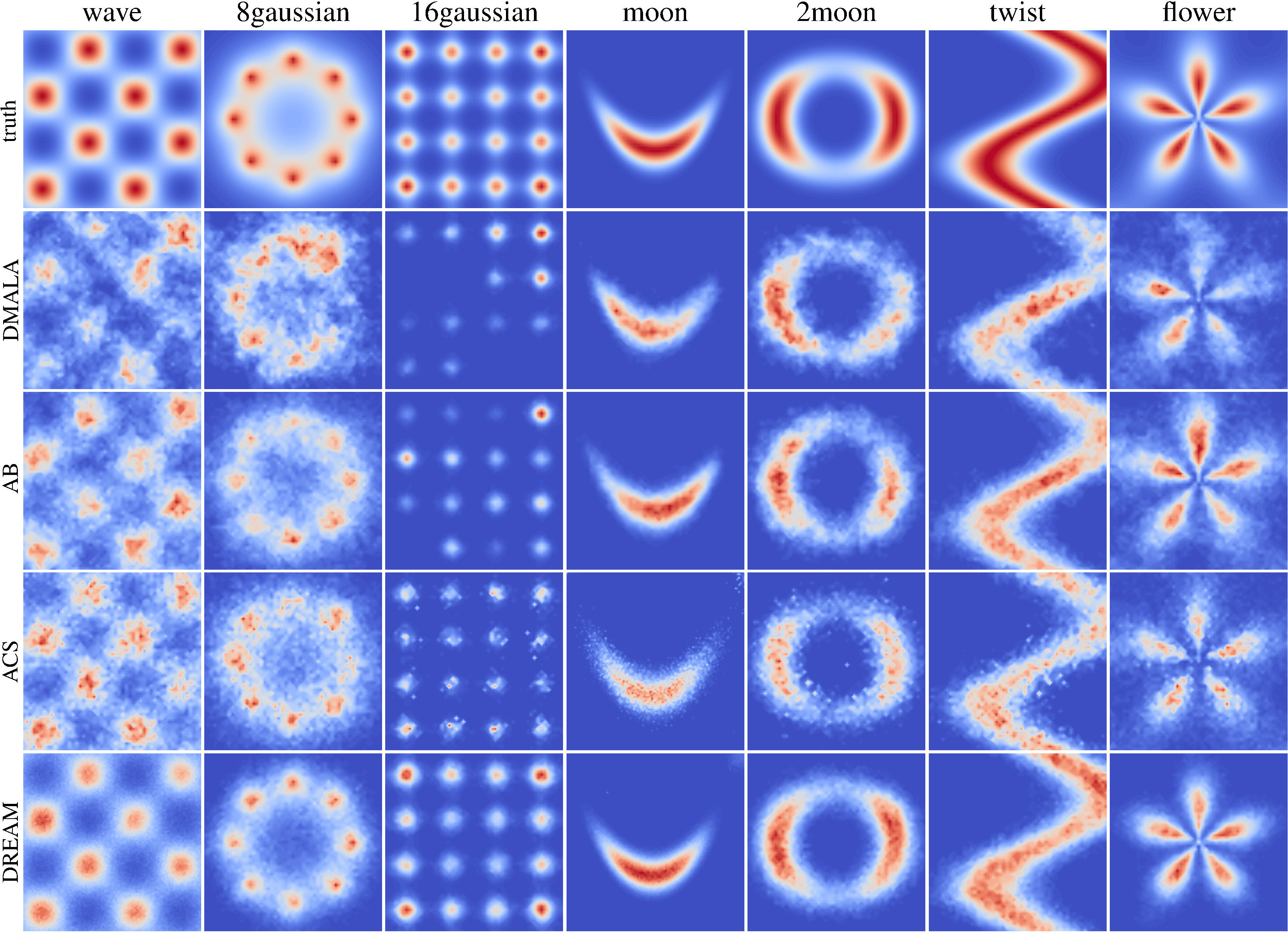}

\caption{
Visualization of the true energy function and the empirical energy function yield by DMALA, AB, ACS, and DREAM. Seven energy functions are tested here, which include wave, eight Gaussians, sixteen Gaussians, moon, two moons, twist, and flower energy functions. Red colors denote high-density regions, and blue colors represent low-density regions.
% \emph{Middle:} Empirical energy distributions sampling from DMALA.
% \emph{Bottom:} Empirical energy distributions sampling from DREAM.
% We defer the visualization of baselines to the Appendix.
}
% \vspace{-3mm}
\label{appendix:fig:synthetic_result}
\end{figure*}

In our experiments, we evaluate different samplers using Kullback-Leibler (KL) divergence and Maximum Mean Discrepancy (MMD).

\textbf{KL divergence} measures the difference between two probability distributions. Given two distributions $\pi$ and $\tilde{\pi}$, the KL divergence is defined as:
$$
D_{KL}(\pi \parallel \tilde{\pi}) = \sum_{\btheta\in\Theta} \pi(\btheta) \log \frac{\pi(\btheta)}{\tilde{\pi}(\btheta)},
$$
where $\pi(\btheta)$ represents the probability of $\btheta$ under the target distribution, and $\tilde{\pi}(\btheta)$ represents the probability of $\btheta$ under the empirical distribution from the samplers. This metric quantifies how much information is lost when $\tilde{\pi}$ is used to approximate ${\pi}$, with lower values indicating better performance.

\textbf{MMD} is a kernel-based test used to compare distributions. It is computed as:
$$
\text{MMD}^2(\pi, \tilde{\pi}) = \mathbb{E}_{\bm x,\bm  x' \sim \pi}[k(\bm x,\bm  x')] + \mathbb{E}_{\bm y, \bm y' \sim \tilde{\pi}}[k(\bm y, \bm y')] - 2 \mathbb{E}_{\bm x \sim \pi, \bm y \sim \tilde{\pi}}[k(\bm x, \bm y)],
$$
where $k(\bm x, \bm y)$ is a positive-definite kernel function. MMD measures the similarity between the empirical distributions of the generated and target samples. 

In practice, however, directly computing MMD is computationally expensive. Therefore, we use an approximation based on Random Fourier Features (RFF) \citep{rahimi2007random}. 

For two distributions $\pi$ and $\tilde{\pi}$, we first map the data samples $X \sim \pi$ and $Y \sim \tilde{\pi}$ to a new feature space using the random Fourier transformation:
$
\phi(X) = \sqrt{\frac{2}{D}} \cos(W X^T + \tilde b),
$
where $W \in\mathbb R^{D\times\mathbf d}$ are random Gaussian variables sampled from $\mathcal{N}\left(0,1/\tilde\sigma^2\right)$, and $\tilde b$ are random uniform variables in the range $[0, 2\pi]$. The parameter $\tilde\sigma$ controls the kernel bandwidth, and $D$ is the number of random features. Once mapped, the empirical mean feature embeddings for $X$ and $Y$ are computed for both distributions $\mu_X = \frac{1}{n} \sum_{i=1}^{n} \phi(X_i), \ \mu_Y = \frac{1}{m} \sum_{i=1}^{m} \phi(Y_i)$. Finally, the MMD is approximated by the squared difference of the mean embeddings:
$$
\text{MMD}^2(\pi, \tilde{\pi}) \approx \left\|\mu_X - \mu_Y\right\|^2.
$$
This approach allows us to efficiently compute the MMD between two distributions using RFFs.

To evaluate the effectiveness of the proposed sampler, we explore its performance on a set of non-convex discrete energy landscapes that vary in complexity and multimodality. We further compare it with baselines such as DMALA, ACS, and AB. Unless specified otherwise, the default temperature for each sampler is set at 1.0. DMALA is implemented with a step size of 0.15.  AB is used with parameters $\sigma=0.10$ and $\alpha=0.50$. For ACS, a cyclical step size scheduler with an initial step size of 0.60 across 10 cycles is applied. DREAM uses small and large step sizes of 0.15 and 0.60 and temperatures of 1.0 and 5.0.

Figure \ref{appendix:fig:synthetic_result} illustrates a comparison between the empirical distributions obtained from different samplers and the ground truth (top row). A clear distinction can be observed between the discrete samplers in terms of their ability to capture the full complexity of the landscape. From the figure, DREAM produces the most balanced and comprehensive empirical distribution, which captures all significant modes of the energy landscape. The improvements are significant in tasks of approximating wave and multi-Gaussian energy functions. While other discrete samplers (such as DMALA, AB, ACS) fail to capture all modes, DREAM exhibits the most comprehensive exploration, as reflected in the uniformity of the empirical distribution across all modes. Its ability to distribute samples effectively, even in the presence of disconnected modes and sharp energy barriers, demonstrates its robustness in navigating complex discrete energy landscapes. By contrast, DMALA shows a heavy concentration of samples around certain modes, which indicates that it struggles to escape local minima. This leads to poor coverage of the landscape and a lack of diversity in the sampled regions. ACS and AB perform better in terms of covering multiple modes but still show uneven sample distributions. Some modes are under-sampled, while others are over-sampled, particularly in regions with shallow energy gradients.

% \clearpage

\subsection{Sampling from Ising Models}

We sampled from the Ising model using multiple samplers with a default temperature of 1.0, unless otherwise specified. DULA utilized a step size of 0.20, while DMALA had a step size of 0.40. For ACS, we employed a cyclical step size scheduler with 10 cycles and an initial step size of 0.30. ACS with MH corrections used an initial step size of 5.0. For DREXEL, small and large step sizes were set at 0.15 and 0.50, with temperatures at 1.0 and 5.0. DREAM followed a similar temperature schedule, with a small step size of 0.35 and a large size of 0.50.

In this study, different samplers are evaluated with log Root Mean Square Error (log RMSE). Log RMSE evaluates prediction accuracy for comparing the true values from $\pi$ and the predictions from $\tilde{\pi}$, which can be adapted as:
$$\text{log RMSE} = \log \left( \sqrt{\frac{1}{n} \sum_{i=1}^{n} (\pi(\bm x_i) - \tilde{\pi}(\bm x_i))^2} \right),$$
where $\pi(\bm x_i)$ is the true value under the target distribution, and $\tilde{\pi}(\bm x_i)$ is the corresponding approximation from the empirical distribution.

% \begin{table*}[ht]
% \centering
% % \resizebox{0.95\textwidth}{!} % Large text
% {
% \begin{tabular}{ccccccc} \toprule
%  Samplers  & $\log$ RMSE  \\ \midrule
% bDREXEL    & -2.084$\pm$0.088 \\
% bDREAM    & -2.081$\pm$0.091   \\ 
% DREXEL &  -4.242$\pm$0.204  \\
% DREAM &  -4.486$\pm$0.190  \\
% \bottomrule 
% \end{tabular}}
% \caption{Results of Ising Model with different samplers (log RMSE).}
% \label{tab:ising_log_rmse}
% \end{table*}

% \begin{table*}[ht]
% \centering
% % \resizebox{0.95\textwidth}{!} % Large text
% {
% \begin{tabular}{ccccccc} \toprule
%  $\sigma^2$ & -5 & 0.0  & 2.0 & 5.0 & 10.0 \\ \midrule
% bDREXEL  & -2.084$\pm$0.088 & -2.090$\pm$0.091 & -2.082$\pm$0.089 & -2.085$\pm$0.093 & -2.003$\pm$0.080 \\
% bDREAM & -2.081$\pm$0.091 & -2.082$\pm$0.080 & -2.091$\pm$0.067 & -2.075$\pm$0.082 & -2.087$\pm$0.095   \\ 
% DREXEL & -4.655$\pm$0.210 & -4.567$\pm$0.260 & -4.426$\pm$0.389 & -3.829$\pm$0.272 & -2.985$\pm$0.556  \\
% DREAM & -4.596$\pm$0.163 & -4.589$\pm$0.158 & -4.593$\pm$0.149 & -4.584$\pm$0.148 & -4.589$\pm$0.151  \\
% \bottomrule 
% \end{tabular}}
% \caption{EBM learning results (log-likelihood) on the test set.}
% \label{tab:ebm}
% \end{table*}

% \clearpage
\subsection{Sampling from Restricted Boltzmann Machines}

We trained RBMs with the Adam optimizer with a learning rate of 0.001, over 1,000 iterations, and a batch size of 128. For training, we used contrastive divergence (CD), which approximates the log-likelihood gradient by performing $k=10$ Gibbs sampling steps. The gradient of the log-likelihood for an RBM is given by:
$$
\nabla_{\btheta} \log P_{\btheta}(\bm x) = \mathbb{E}_{P_{\text{data}}}[\nabla_{\btheta} \log P_{\btheta}(\bm x)] - \mathbb{E}_{P_{\btheta}}[\nabla_{\btheta} \log P_{\btheta}(\bm x)],
$$
where $\bm x$ is the visible layer, and $\btheta$ are the model parameters. The first term corresponds to the data distribution, while the second term is the expectation under the model’s distribution. It should be noted that direct computation of the model distribution expectation is expensive, and CD approximates the second term by running $k$-step Gibbs sampling to obtain samples from the model. This approximation enables efficient training of RBMs by focusing on the contrast between the observed and modeled distributions.

To sample from the RBMs, we employed several discrete samplers at a default temperature of 1.0 unless otherwise noted. DMALA used a step size of 0.15, and ACS applied a cyclical step size scheduler with 10 cycles, starting at a step size of 0.50. The Any-scale Balanced (AB) sampler was configured with $\sigma=0.10$ and $\alpha=0.50$. For DREAM, the small and large step sizes were set between 0.15–0.20 and 0.40–0.50, with temperatures set at 1.0 and 2.0.

\begin{table}[!htbp]
\centering\caption{Hyperparameters used in sampling RBM models. For each dataset, 
% the number of Gibbs sampling steps (cd), 
low-temperature chain step size ($\alpha_1$), high-temperature chain step size ($\alpha_2$), and high temperatures ($\tau_2$) are recorded.}\label{tab:rbm_parameters}
\begin{tabular}{ccccccc}
\toprule
Dataset  & $\alpha_1$ & $\alpha_2$ & $\tau_2$ & log MMDs \\ \midrule
MNIST    & 0.21      & 0.32      & 7.15  & {-6.349}$\pm\scriptstyle {0.061}$           \\
eMNIST   & 0.21      & 0.42      & 5.92 &  {-6.148}$\pm\scriptstyle {0.077}$           \\
kMNIST   & 0.15      & 0.44      & 4.84  &  {-5.875}$\pm\scriptstyle {0.040}$          \\
Fashion  & 0.23      & 0.34      & 7.50  &  {-5.901}$\pm\scriptstyle {0.077}$          \\
Omniglot & 0.23      & 0.35      & 7.96  &   {-6.626}$\pm\scriptstyle {0.057}$         \\
Caltech  & 0.25      & 0.38      & 5.93 &    {-5.810}$\pm\scriptstyle {0.085}$        \\ \bottomrule
\end{tabular}
\end{table}
\subsection{Learning Deep Energy-Based Models}
We trained Deep EBMs using a ResNet-64 backbone and optimized the model with the Adam optimizer at a fixed learning rate of 0.001 without gradient clipping. The model was trained for 50,000 iterations with a batch size of 256. we employed Persistent Contrastive Divergence (PCD) to approximate the intractable likelihood gradient, which builds upon standard contrastive divergence by maintaining persistent Markov chains throughout training. It allows for more stable and accurate sampling. Specifically, the model's log-likelihood gradient is given by:
$$
\nabla_{\btheta} \log P_{\btheta}(\bm x) = \mathbb{E}_{P_{\text{data}}}[\nabla_{\btheta} \log P_{\btheta}(\bm x)] - \mathbb{E}_{P_{\btheta}}[\nabla_{\btheta} \log P_{\btheta}(\bm x)],
$$
where the second term (the model expectation) is intractable. PCD approximates this by updating samples across training iterations via Gibbs sampling, which ensures that the Markov chain does not restart after each parameter update. Additionally, a replay buffer containing 1,000 past samples is used to further stabilize training. The buffer stores past model samples and reuses them to reduce variance, thus improving both the efficiency and stability of the learning process.

To evaluate Deep EBMs, we applied Annealed Importance Sampling (AIS) with DULA to estimate the test log-likelihoods. AIS is a technique used to estimate partition functions by smoothly interpolating between a known distribution and the target distribution. This is achieved by introducing a sequence of intermediate distributions:
$$
P_t(\bm x) = \frac{1}{Z_t} \exp(-\beta_t E(\bm x)),
$$
where $t$ denotes the current annealing step, $E(\bm x)$ is the energy function, $Z_t$ is the partition function, and $\beta_t$ is a temperature that gradually transitions between 0 and 1 over the course of the annealing process. When $\beta_t = 0$, the intermediate distribution is identical to the proposal distribution (which we can sample from easily). When $\beta_t = 1$, the intermediate distribution becomes the target distribution, which is more complex and generally intractable to sample directly. To adjust $\beta_t$, we typically choose a monotonic schedule that increases smoothly from 0 to 1 over the course of the AIS process. A common choice is a linear interpolation ($\beta_t = t/T, \ t = 0, 1, 2, \dots, T$) or an exponential schedule ($\beta_t = \left( t/T \right)^2$), where $\beta_t$ increases evenly across $T$ annealing steps. AIS computes an estimate of the partition function by sampling from these intermediate distributions and adjusting the importance weights over time:
$$
\tilde{Z}_\btheta = Z_0 \prod_{t=1}^T \frac{P_{t}(\bm x)}{P_{t-1}(\bm x)},
$$
where $Z_0$ is an initial distribution that is easy to sample from, which serves as a starting point for the annealing process. Typically, $Z_0$ is chosen to be the partition function of a simple proposal distribution $p_0(x)$, which is often a uniform distribution or a Gaussian distribution with parameters that are easy to compute. In our experiments, we used AIS with 40 samples and 30,000 annealing steps. DULA was configured with a step size of 0.08 and a temperature of 1.00. Detailed hyperparameters for training Deep EBMs are listed in Table \ref{tab:ebm:hyper}.

% \paragraph{Hyper parameters}

\begin{table*}[!htbp]
\centering
\caption{Hyperparameters used in learning Deep EBMs. From top to bottom, hyperparameters in Static MNIST, Dynamic MNIST, Omniglot, and Caltech Silhouettes are recorded.}
\label{tab:ebm:hyper}
\resizebox{0.9\textwidth}{!}{
\begin{tabular}{lcccccc} \toprule
Static MNIST      & DULA & DMALA & bDREXEL & bDREAM & DREXEL & DREAM \\ \midrule
\multirow{2}{*}{Step size}         & 0.08 & 0.10  & 0.05       & 0.05        & 0.11      & 0.10       \\
 & -    & -     & 0.15       & 0.15        & 0.25      & 0.30       \\ \cmidrule{2-7}
\multirow{2}{*}{Temperature}              & 1.0  & 1.0   & 1.0        & 1.0         & 1.0       & 1.0        \\
   & -    & -     & 5.0        & 5.0         & 5.0       & 5.0        \\ \cmidrule{2-7}
Correction        & -  &  -  &  0.00   & 0.00  &  0.00      & 0.00  \\ \bottomrule 
\\ \toprule
Dynamic MNIST      & DULA & DMALA & bDREXEL & bDREAM & DREXEL & DREAM \\ \midrule
\multirow{2}{*}{Step size}         & 0.08 & 0.10  & 0.05       & 0.05        & 0.11      & 0.11       \\
 & -    & -     & 0.15       & 0.15        & 0.25      & 0.25       \\ \cmidrule{2-7}
\multirow{2}{*}{Temperature}     & 1.0  & 1.0   & 1.0        & 1.0         & 1.0       & 1.0        \\
   & -    & -     & 5.0        & 5.0         & 5.0       & 5.0        \\ \cmidrule{2-7}
Correction        & -  &  -  &  0.00   & 0.00  &  0.00      & 0.00  \\ \bottomrule 
\\ \toprule
Omniglot      & DULA & DMALA & bDREXEL & bDREAM & DREXEL & DREAM \\ \midrule
\multirow{2}{*}{Step size}    & 0.08 & 0.10  & 0.05       & 0.05        & 0.08      & 0.08       \\
 & -    & -     & 0.15       & 0.15        & 0.15      & 0.15       \\ \cmidrule{2-7}
\multirow{2}{*}{Temperature}              & 1.0  & 1.0   & 1.0        & 1.0         & 1.0       & 1.0        \\
   & -    & -     & 5.0        & 5.0         & 5.0       & 5.0        \\ \cmidrule{2-7}
Correction        & -  &  -  &  0.00   & 0.00  &  1.00      & 0.00  \\ \bottomrule 
\\ \toprule
Caltech      & DULA & DMALA & bDREXEL & bDREAM & DREXEL & DREAM \\ \midrule
\multirow{2}{*}{Step size}         & 0.08 & 0.10  & 0.05       & 0.05        & 0.08      & 0.08       \\
 & -    & -     & 0.15       & 0.15        & 0.20      & 0.20       \\ \cmidrule{2-7}
\multirow{2}{*}{Temperature}              & 1.0  & 1.0   & 1.0        & 1.0         & 1.0       & 1.0        \\
   & -    & -     & 5.0        & 5.0         & 5.0       & 5.0        \\ \cmidrule{2-7}
Correction        & -  &  -  &  0.00   & 0.00  &  0.00      & 0.00  \\ \bottomrule 
\end{tabular}}
\end{table*}

For a consistent comparison with previous works \citep{zhang2022langevin}, we record its log-likelihood on the test set after 50,000 iterations. In general, Table \ref{tab:ebm:50k} yields consistently lower log-likelihood than the results in Table \ref{tab:ebm:20k} since they train with more iterations. But similar trends are shown in Table \ref{tab:ebm:20k} as well: DREAM consistently achieved the highest log-likelihoods across all datasets, with significant improvements on Omniglot and Caltech Silhouettes. This demonstrates that the proposed swap mechanism in \eqref{eq:swap_history} effectively corrects the imbalance, which leads to improved log-likelihood estimates across diverse image datasets. For the MNIST datasets, both DREXEL and DREAM showed competitive performance. bDREXEL and bDREAM generally perform worse than DREAM, with consistently lower log-likelihoods, which further confirms the advantage of historical energy corrections. These findings suggest that incorporating MH steps is crucial for enhancing performance in discrete sampling tasks. DULA and DMALA exhibit the lowest performance overall, which emphasizes the benefits of MH steps and the need to consider DREXEL and DREAM to enhance exploration. 

% \subsection{EBM Results}\label{appendix:subsec:ebm}
\begin{table*}[!htbp]
\centering
\caption{EBM learning results (log-likelihood) on the test set after 50,000 iterations.}
\label{tab:ebm:50k}
\resizebox{0.999\textwidth}{!}{
\begin{tabular}{lcccccc} \toprule
Dataset             &  DULA   & DMALA   & bDREXEL & bDREAM & DREXEL & DREAM \\ \midrule
Static MNIST        & $-79.672\scriptstyle \pm 0.882$ & $-77.581\scriptstyle \pm 0.518$ & $-77.212\scriptstyle \pm 1.348$ & $-76.840\scriptstyle \pm 1.219$ & $-75.685\scriptstyle \pm 0.854$  & $-74.883\scriptstyle \pm 0.893$   \\
Dynamic MNIST       & $-81.144\scriptstyle \pm 0.321$ & $-79.411\scriptstyle \pm 0.593$ & $-81.273\scriptstyle \pm 1.578$ & $-81.043\scriptstyle \pm 1.054$ & $-71.091\scriptstyle \pm 0.612$  & $-70.905\scriptstyle \pm 0.579$   \\
Omniglot            &$-114.203\scriptstyle \pm 6.465$ &$-109.09\scriptstyle \pm 2.564$ & $-94.382\scriptstyle \pm 2.575$ & $-90.807\scriptstyle \pm 2.431$ & $-89.971\scriptstyle \pm 2.057$  & $-89.643\scriptstyle \pm 2.153$   \\
Caltech Silhouettes & $-102.546\scriptstyle \pm 5.189$ &  $-98.554\scriptstyle \pm 4.049$ & $-96.073\scriptstyle \pm 5.651$ & $-93.969\scriptstyle \pm 5.153$ & $-89.764\scriptstyle \pm 3.412$   & $-86.624\scriptstyle \pm 3.451$   \\  \bottomrule 
\end{tabular}}
\end{table*}

\begin{figure*}[!htbp]
\centering
% \subfigure[Acceptance rate.]{
\includegraphics[width=2.5 in]{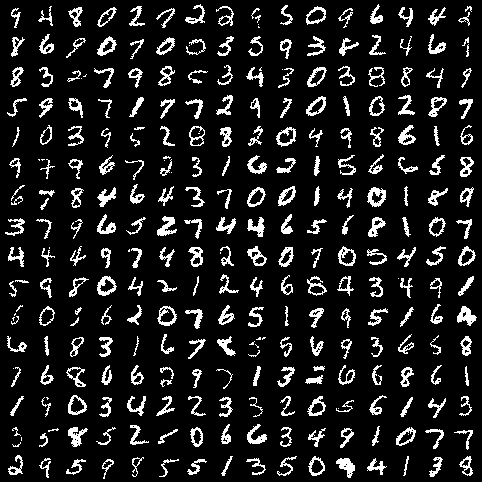}\label{fig:ebm:static_mnist}
% }
% \subfigure[Swap rate.]{
\includegraphics[width=2.5 in]{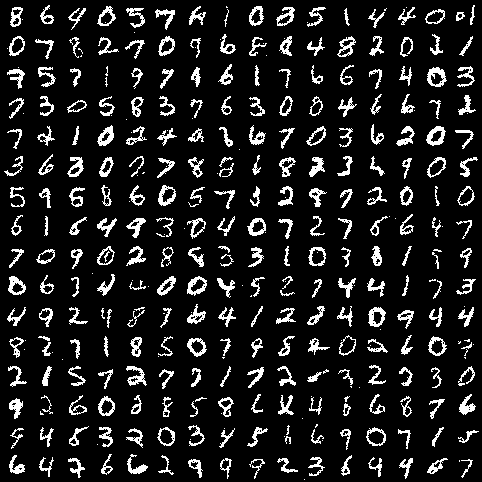}\label{fig:ebm:dynamic_mnist}
% }
% \subfigure[Log RMSE over iterations.]{
\includegraphics[width=2.5 in]{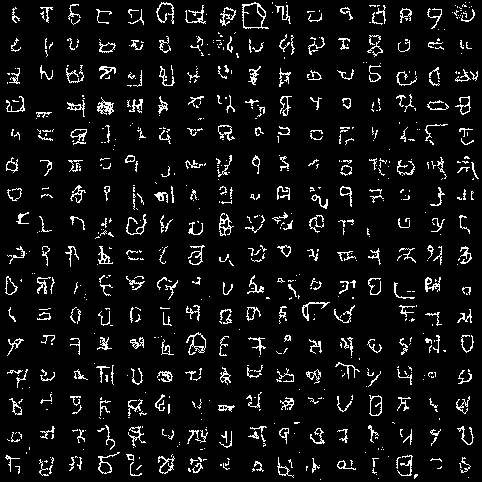}\label{fig:ebm:omniglot}
% }
% \subfigure[Log RMSE over runtime.]{
\includegraphics[width=2.5 in]{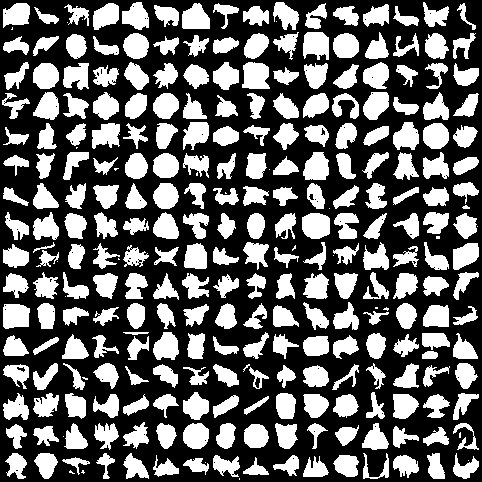}\label{fig:ebm:catech}
% }
\caption{Deep RBMs sampling results from DREAM. \textbf{Top Left:} Static MNIST; \textbf{Top Right:} Dynamic MNIST; \textbf{Bottom Left:} Omniglot; \textbf{Bottom Right:} Caltech Silhouettes.}\label{fig:ebm_sampling}
\end{figure*}

Here we provide the generated results (Figure \ref{fig:ebm_sampling}) from DREAM across Static MNIST, Dynamic MNIST, Omniglot, and Caltech Silhouettes. These images demonstrate the ability of trained deep EBMs to capture the underlying data distribution. The deep EBM is capable of producing high-quality samples that visually resemble the training data, which indicates that the learned energy function effectively models the complex, high-dimensional structure of the data.

\end{document}